\documentclass[preprint]{elsarticle}
\usepackage{amsmath,amssymb,bm,graphicx,color}
\usepackage{lineno}
\usepackage{colortbl}
\usepackage[colorlinks, linkcolor=red, anchorcolor=blue, citecolor=green]{hyperref}

\usepackage[titletoc]{appendix}
\usepackage{bbm}

\usepackage{amsthm}

\newcommand{\mbf}[1]{\mathbf{#1}}

\definecolor{mygray}{gray}{.9}
\definecolor{lightgray}{gray}{.7}
\usepackage{multirow}
\usepackage{subfigure}

\journal{}

\begin{document}
	
	\begin{frontmatter}
		
		\title{Scalable Multi-Task Gaussian Processes with Neural Embedding of Coregionalization}
		
\author[address1]{Haitao Liu}
\ead{htliu@dlut.edu.cn}

\author[address1]{Jiaqi Ding}
\ead{jqding@mail.dlut.edu.cn}

\author[address1]{Xinyu Xie}
\ead{xyxie@mail.dlut.edu.cn}

\author[address2]{Xiaomo Jiang}
\ead{xiaomojiang2019@dlut.edu.cn}

\author[address1]{Yusong Zhao}
\ead{yszhao@mail.dlut.edu.cn}

\author[address1]{Xiaofang Wang\corref{mycorrespondingauthor}}
\ead{dlwxf@dlut.edu.cn}

\cortext[mycorrespondingauthor]{Corresponding author}
\address[address1]{School of Energy and Power Engineering, Dalian University of Technology, China, 116024}
\address[address2]{Digital Twin Laboratory for Industrial Equipment, Dalian University of Technology, China, 116024.}

\begin{abstract}
Multi-task regression attempts to exploit the task similarity in order to achieve knowledge transfer across related tasks for performance improvement. The application of Gaussian process (GP) in this scenario yields the non-parametric yet informative Bayesian multi-task regression paradigm. Multi-task GP (MTGP) provides not only the prediction mean but also the associated prediction variance to quantify uncertainty, thus gaining popularity in various scenarios. The linear model of coregionalization (LMC) is a well-known MTGP paradigm which exploits the dependency of tasks through linear combination of several independent and diverse GPs. The LMC however suffers from high model complexity and limited model capability when handling complicated multi-task cases. To this end, we develop the neural embedding of coregionalization that transforms the latent GPs into a high-dimensional latent space to induce rich yet diverse behaviors. Furthermore, we use advanced variational inference as well as sparse approximation to devise a tight and compact evidence lower bound (ELBO) for higher quality of scalable model inference. Extensive numerical experiments have been conducted to verify the higher prediction quality and better generalization of our model, named NSVLMC, on various real-world multi-task datasets and the cross-fluid modeling of unsteady fluidized bed.
\end{abstract}

\begin{keyword}
Multi-task Gaussian process \sep Linear model of coregionalization \sep Neural embedding \sep Diversity \sep Tighter ELBO
\end{keyword}

\end{frontmatter}


\section{Introduction}
In comparison to the conventional single-task learning, multi-task learning (MTL)~\cite{zhang2021a} provides a new learning paradigm to leverage knowledge across related tasks for improving the generalization performance of tasks. The community of multi-task learning has overlaps with other domains like transfer learning~\cite{zhuang2021atl}, multi-view learning~\cite{sun2013a} and multi-fidelity modeling~\cite{fernndez-godino2016review}. Among current MTL paradigms, multi-task Gaussian process (MTGP), the topic of this paper, inherits the non-parametric, Bayesian property of Gaussian process (GP)~\cite{williams2006gaussian} to have not only the prediction mean but also the associated prediction variance, thus showcasing widespread applications, e.g., multi-task regression and classification, multi-variate time series analysis~\cite{drichen2015multitask}, multi-task Bayesian optimization~\cite{swersky2013multi, kandasamy2019multi}, and multi-view learning~\cite{mao2021multiview}.

To date, various MTGP models have been proposed in literature. Among them, the linear model of coreionalization (LMC)~\cite{goovaerts1997geostatistics, bonilla2007multi, teh2005semiparametric} is a well-know MTGP framework that linearly mixes $Q$ independent, latent GPs for modeling $C$ related tasks simultaneously. The latent GPs in LMC achieve knowledge transfer since they are shared across tasks, while the task-related mixing coefficients adapt the behaviors for specific tasks. Other popular MTGPs include for example the convolved GP~\cite{alvarez2008sparse}, Co-Kriging~\cite{myers1982matrix}, and stacked GP~\cite{neumann2009stacked}. For the details of various MTGPs, readers are suggested to refer to these surveys, reports and implementations~\cite{alvarez2012kernels, liu2018remarks, brevault2020overview, wolff2021mogptk}. It is notable that this paper mainly focuses on the LMC-type MTGPs.

The improvements over the original LMC model mainly raise from two views. The first is improving the capability of multi-task modeling. This can be done by simply increasing the number $Q$ of latent GPs, which however significantly increases the model complexity, making it unaffordable on large-scale datasets. Alternatively, instead of using the conventional stationary kernels, like the squared exponential (SE) kernel in~\eqref{eq_SE}, we could devise the more expressive spectral mixing kernel which could take for example the phase shift and decays between tasks into account, thus enhancing the learning of complicated cross-task relationships~\cite{parra2017spectral, chen2019multioutput}. As for task similarity, traditional LMC adopts global, constant coefficients to mix the independent latent GPs. As an improvement, we could employ additional GPs for modeling complicated, input-varying task correlations, see~\cite{wilson2012gaussian}. Besides, various likelihood distributions, for example, the student-$t$ and probit distributions, have been utilized to extend MTGPs for different downstream scenarios, for example, classification and heterogeneous modeling~\cite{chen2020multivariate, moreno-muoz2019continual, moreno-muoz2018heterogeneous}. Finally, as for the improvement of model structure, we could employ the residual components to account for negative transfer~\cite{nguyen2014collaborative, liu2018cope}, the auto-regressive modeling to transfer the knowledge of previous tasks sequentially~\cite{requeima2019gaussian, perdikaris2017nonlinear}, or the combination of powerful deep models to enhance the representational learning~\cite{kandemir2015asymmetric, jankowiak2019neural, mao2021multiview}.

The second is improving the scalability of LMC for tackling massive data, which is an urgent demand for MTL due to the simultaneous modeling of multiple tasks. To this end, we usually leverage the idea from scalable GPs that have been recently compared and reviewed in~\cite{liu2019understanding, liu2020gaussian}. The majority of scalable LMCs relies on the framework of sparse approximation~\cite{snelson2006sparse, titsias2009variational, hensman2013gaussian}, which introduces $M$ inducing variables to be the sufficient statistics of $N$ latent function values for a task with $M \ll N$, thus greatly reducing the cubic model complexity~\cite{alvarez2011computationally, nguyen2014collaborative, NIPS2015_3b3dbaf6, ashman2020sparse, bruinsma2020scalable}. Other complexity reduction strategies have also been investigated through for example the distributed learning~\cite{chiplunkar2016approximate}, the natural gradient assisted stochastic variational inference~\cite{giraldo2021a}, and the exploitation of Kronecker structure in kernel matrix~\cite{stegle2011efficient, rakitsch2013all}. Recently, the efficient and effective modeling of many outputs, i.e., ``big $C$'', has been investigated through for example the manifold learning and the tensor decomposition~\cite{perdikaris2016multifidelity, yu2017tensor, zhe2019scalable, wang2020multi}.

To further improve the model capability as well as the scalability of LMC, this paper devises a new paradigm, named NSVLMC, to flexibly enhance the number and diversity of latent GPs while keeping the automatic regularization of GP and the desirable model complexity. The main contributions of our work are three-folds:
\begin{itemize}
\item We propose to use the flexible and powerful \textit{neural embedding} to transform the latent independent GPs into a \textit{higher-dimensional} yet \textit{diverse} space, thus greatly improving the model expressivity of LMC;
\item Under the framework of sparse approximation, we further derive \textit{tighter} and \textit{more compacted} evidence lower bound (ELBO) in order to enhance the model inference quality of scalable LMC when handling massive multi-task data;
\item We finally conduct comprehensive experiments to investigate and verify the methodological characteristics and superiority of the proposed model over existing MTGPs.
\end{itemize}

The remaining of this paper is organized as follows. Section~\ref{sec_gp_mtgp} first has a brief introduction of GP and LMC. Thereafter, section~\ref{sec_nlmc} introduces the proposed NSVLMC model, followed by the scalable variational inference for efficient and effective model training in section~\ref{sec_vi}. Then, section~\ref{sec_exp} conducts extensive numerical experiments on various multi-tasks scenarios in order to verify the superiority of our NSVLMC model. Finally, section~\ref{sec_conclusion} provides concluding remarks.

\section{Preliminaries} \label{sec_gp_mtgp}
\subsection{Gaussian process}
For a single-task regression task, the data-driven GP learns the mapping between the input domain $\mathcal{X} \in \mathbb{R}^D$ and the output domain $\mathcal{Y} \in \mathbb{R}$ through the following expression
\begin{align}
y(\mbf{x}) = f(\mbf{x}) + \epsilon,
\end{align}
where $\epsilon \sim \mathcal{N}(\epsilon|0, \nu_{\epsilon})$ is an independent and identically distributed (\textit{i.i.d.}) noise; and the latent function $f(.): \mathbb{R}^D \mapsto \mathbb{R}$ defined over the functional space is placed with a GP prior
\begin{align}
f(\mbf{x}) \sim \mathcal{GP}(m(\mbf{x}), k(\mbf{x}, \mbf{x}') ),
\end{align}
where $m(\mbf{x})$ is the mean function which usually takes zero without loss of generality, and $k(\mbf{x}, \mbf{x}')$ is the kernel (covariance) function describing the similarity between two arbitrary data points, with the popular choice being the squared exponential (SE) kernel equipped with automatic relevance determination (ARD) as
\begin{align} \label{eq_SE}
k(\mbf{x}, \mbf{x}') = \sigma_f^2 \exp\left(-\frac{1}{2} \sum_{i=1}^D \frac{(x_i - x'_i)^2}{l_i^2} \right),
\end{align}
where the hyperparameter $\sigma_f^2$ is an output scale, and $l_i$ is the length-scale representing the variation or smoothness along the $i$-th input dimension.\footnote{For other commonly used kernels, e.g., the Mat\'{e}rn family, please refer to~\cite{williams2006gaussian}.}

After observing $N$ training data $\mathcal{D} = \{\mbf{X} \in \mathbb{R}^{N \times D}, \mbf{y} \in \mathbb{R}^N \}$, we optimize the hyperparameters of GP by maximizing the marginal likelihood, which achieves automatic regularization, as
\begin{align}
p(\mbf{y}) = \mathbb{E}_{p(\mbf{f})} [p(\mbf{y}|\mbf{f})] = \mathcal{N}(\mbf{y}|\mbf{0}, \mbf{K} + \nu_{\epsilon} \mbf{I}),
\end{align}
where the kernel matrix $\mbf{K} = k(\mbf{X}, \mbf{X}) \in \mathbb{R}^{N \times N}$, the determinant or inversion of which is an $\mathcal{O}(N^3)$ operator; and the multi-variate Gaussian prior $p(\mbf{f}) = \mathcal{N}(\mbf{f}|\mbf{0}, \mbf{K})$. 

Given the data as well as the optimized hyperparameters, we update our Gaussian beliefs and  devise the Gaussian posterior through the Bayes' rule $p(\mbf{f}|\mbf{y}) \propto p(\mbf{y}|\mbf{f}) p(\mbf{f})$, which is thereby utilized to perform Gaussian prediction at a test point $\mbf{x}_*$ as $p(y_*|\mbf{y}) = \mathcal{N}(y_*|\mu_*, \nu_*)$, with the mean and variance expressed respectively as
\begin{align}
\mu_* =& \mbf{k}_*^{\mathsf{T}} [\mbf{K} + \nu_{\epsilon} \mbf{I}]^{-1} \mbf{y}, \\
\nu_* =& k_{**} - \mbf{k}_*^{\mathsf{T}} [\mbf{K} + \nu_{\epsilon} \mbf{I}]^{-1} \mbf{k}_* + \nu_{\epsilon},
\end{align}
where the vector $\mbf{k}_* = k(\mbf{X}, \mbf{x}_*) \in \mathbb{R}^N$.

\subsection{Linear model of coregionalization}
\begin{figure}[t!]
	\centering
	\includegraphics[width=1.0\textwidth]{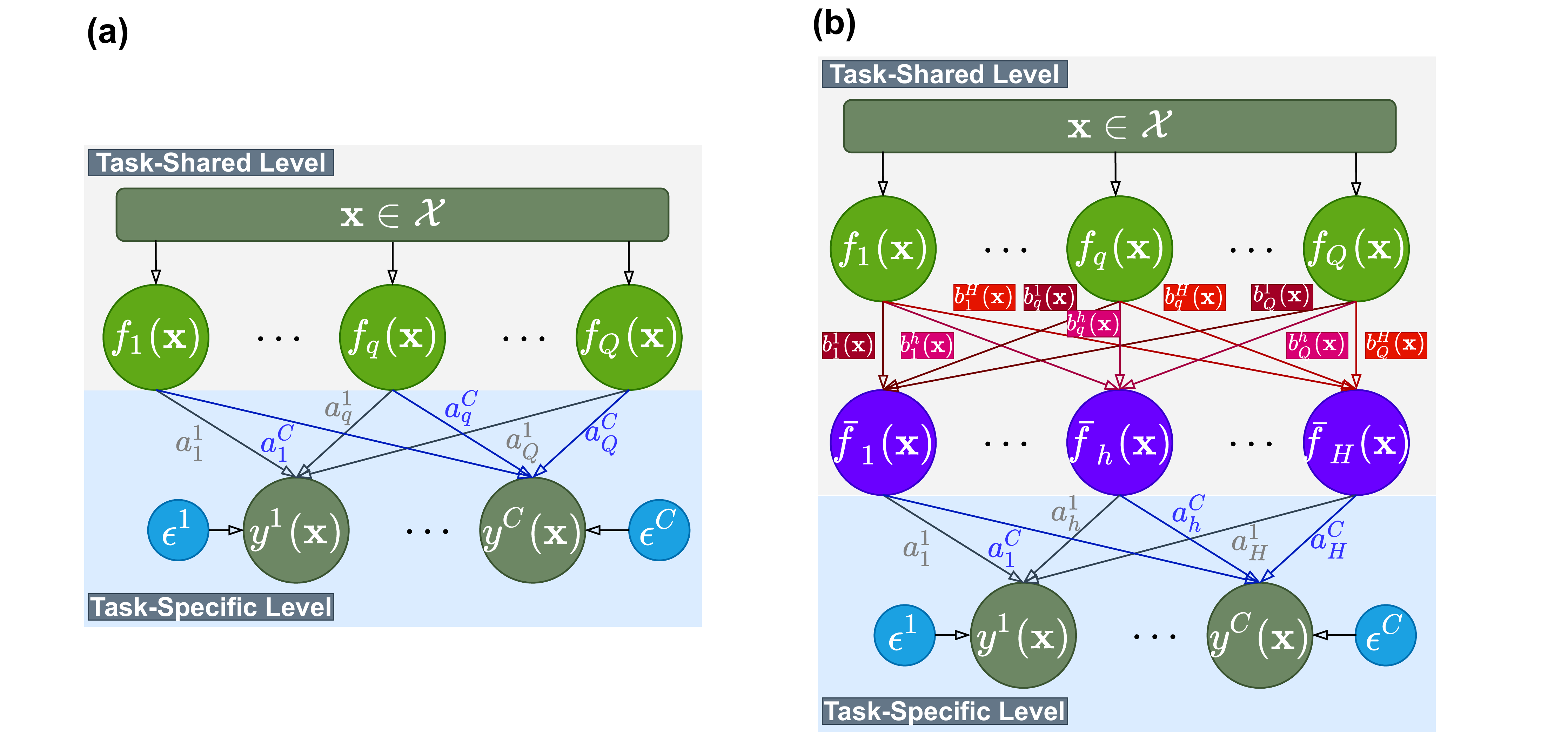}
	\caption{Graphical models of (a) the original LMC and (b) the proposed model for multi-task learning.}
	\label{fig_LMCvsNLMC}
\end{figure}

For the joint learning of $C$ related regression tasks, the multi-task GP should be enabled to measure the similarity of tasks in order to enhance knowledge transfer across tasks, which distinguishes it from conventional single-task GP. As shown in Fig.~\ref{fig_LMCvsNLMC}(a), we consider the following linear model of coregionalization (LCM) for the $c$-th task defined in the input space $\mathcal{X}$ as
\begin{align} \label{eq_mogp}
y^c(\mbf{x}) = \sum_{q=1}^Q a^c_q f_q(\mbf{x})+ \epsilon^c,
\end{align}
where the $Q$ latent functions $\{f_q \sim \mathcal{GP}(0, k_q(.,.)) \}_{q=1}^Q$ are \textit{independent} and \textit{diverse} GPs which however are \textit{shared} across $C$ tasks.\footnote{These independent GPs can have their own kernels $\{k_q(.,.) \}_{q=1}^Q$ or, for simplicity, share the same kernel $k(.,.)$.} Note that the independency of latent GPs simplifies model inference, while the diversity enhances multi-scale feature extraction of related tasks; the task similarity (knowledge transfer) is achieved through using the task-specific coefficients $\{a_q^c\}_{1\le q\le Q}^{1 \le c \le C}$ to linearly mix the shared latent GPs; and finally, $\epsilon^c = \mathcal{N}(\epsilon^c|0,\nu^c_{\epsilon})$ is the \textit{i.i.d.} noise for the $c$-th task, which remains possible knowledge transfer even when the tasks have the same training inputs~\cite{bonilla2007multi, rakitsch2013all}. Other similar and extended expressions for LMC can be found in~\cite{alvarez2012kernels}.

The vector-valued form of model~\eqref{eq_mogp} is
\begin{align} \label{eq_lmc_matrix}
\mbf{y}(\mbf{x}) = \mbf{A} \mbf{f}(\mbf{x}) + \bm{\epsilon},
\end{align}
where the $C$ observations at point $\mbf{x}$ are $\mbf{y}(\mbf{x}) = [y^1(\mbf{x}), \cdots, y^C(\mbf{x})]^{\mathsf{T}}$, the $Q$ shared latent function values $\mbf{f}(\mbf{x}) = [f_1(\mbf{x}), \cdots, f_Q(\mbf{x})]^{\mathsf{T}}$, and finally, the similarity matrix (also known as coregionalization matrix) $\mbf{A} = [\mbf{a}^1, \cdots, \mbf{a}^C]^{\mathsf{T}} \in \mathbb{R}^{C \times Q}$ where $\mbf{a}^C = [a_1^C, \cdots, a_Q^C]^{\mathsf{T}}$.

Without loss of generality, suppose that we have the \textit{heterotopic} training inputs $\mbf{X} = \{\mbf{X}^c \in \mathbb{R}^{N^c \times D}\}_{c=1}^C$ for $C$ related tasks, and the related observations $\mbf{y} = \{\mbf{y}^c \in \mathbb{R}^{N^c} \}_{c=1}^C$. We further define the notations: the total training size $N = \sum_c^C N^c$  for all the tasks, the $q$-th latent function values $\mbf{f}_q^c = [f_q(\mbf{x}^c_1), \cdots, f_q(\mbf{x}^c_{N^c})]^{\mathsf{T}}$ at training points for the $c$-th task, the $q$-th latent function values $\mbf{f}_q =\{ \mbf{f}^c_q\}_{c=1}^C$ for all the tasks, and finally, the overall latent function value set $\mbf{f}=\{\mbf{f}_q \in \mathbb{R}^N \}_{q=1}^Q$. Thereafter, we define the following Gaussian likelihood factorized over both data points and tasks as
\begin{align}
p(\mbf{y}|\mbf{f}) = \prod_{c=1}^C \prod_{i=1}^{N^c} \mathcal{N}\left(y_i^c \left| \sum_{q=1}^Q a_q^c f_q(\mbf{x}_i^c), \nu^c_{\epsilon} \right.\right),
\end{align}
Due to the independency assumption, we also have the following factorized GP priors
\begin{align}
p(\mbf{f}) = \prod_{q=1}^Q p(\mbf{f}_q) = \prod_{q=1}^Q \mathcal{N}(\mbf{f}_q| \mbf{0}, \mbf{K}_q),
\end{align}
where the covariance matrix $\mbf{K}_q = [\mbf{K}_q^{cc'} = k_q(\mbf{X}^c, \mbf{X}^{c'}) \in \mathbb{R}^{N^c \times N^{c'}}]_{1\le c,c' \le C} \in \mathbb{R}^{N \times N}$. Thereafter, similar to the single-task GP, we have the marginal likelihood (model evidence)
\begin{align}
p(\mbf{y}) = \mathcal{N} \left(\mbf{y} \left| \mbf{0}, \sum_{q=1}^Q \bar{\mbf{K}}_q + \bm{\Xi} \right.\right),
\end{align}
where the covariance $\bar{\mbf{K}}_q = [\bar{\mbf{K}}_q^{cc'} = a_q^c a_q^{c'}k_q(\mbf{X}^c, \mbf{X}^{c'})\in \mathbb{R}^{N^c \times N^{c'}}]_{1\le c,c' \le C} \in \mathbb{R}^{N \times N}$ is the scaled version of $\mbf{K}_q$ by taking into account the mixture coefficients, and the diagonal noise matrix $\bm{\Xi} = \mathrm{diag}[\bm{\Xi}^1, \cdots, \bm{\Xi}^C] \in \mathbb{R}^{N \times N}$ with the $c$-th diagonal noise matrix $\bm{\Xi}^c = \nu^c_{\epsilon} \mbf{I}_{N^c}\in \mathbb{R}^{N^c \times N^c}$.

Given the training data and the optimized hyperparameters, we perform predictions akin to GP for the $C$ tasks jointly at an unseen point $\mbf{x}_*$ as $p(\mbf{y}_*|\mbf{y}) = \mathcal{N}(\mbf{y}_*|\bm{\mu}_* \in \mathbb{R}^C, \bm{\Sigma}_* \in \mathbb{R}^{C\times C})$, where the mean and covariance are respectively expressed as
\begin{align}
	\bm{\mu}_* =& \bar{\mbf{K}}_*^{\mathsf{T}} [\bar{\mbf{K}} + \bm{\Xi}]^{-1} \mbf{y}, \\
	\bm{\Sigma}_* =& \bar{\mbf{K}}_{**} - \bar{\mbf{K}}_*^{\mathsf{T}} [\bar{\mbf{K}} + \bm{\Xi}]^{-1} \bar{\mbf{K}}_* + \mathrm{diag}[\bm{\xi}],
\end{align}
where the full covariance matrix $\bar{\mbf{K}} = \sum_{q=1}^Q \bar{\mbf{K}}_q \in \mathbb{R}^{N\times N}$; the covariance matrix $\bar{\mbf{K}_*} = \sum_{q=1}^Q \bar{\mbf{K}}_{q*} = \sum_{q=1}^Q [a_q^c a_q^{c'}k_q(\mbf{X}^c, \mbf{x}_*)]_{1\le c,c' \le C} \in \mathbb{R}^{N\times C}$ describes the correlations between the training and testing data of $C$ tasks; the covariance matrix $\bar{\mbf{K}}_{**} = [a_q^c a_q^{c'}k_q(\mbf{x}_*, \mbf{x}_*)]_{1\le c,c' \le C} \in \mathbb{R}^{C\times C}$ measures the correlations of testing data of $C$ tasks; and finally, the vector $\bm{\xi} = [\nu_{\epsilon}^1, \cdots, \nu_{\epsilon}^C]^{\mathsf{T}}$ collects the independent noise variances of $C$ tasks.

\section{The LMC with neural embedding} \label{sec_nlmc}
It is known that the capability of LMC increases with $Q$ (i.e., the number of latent GPs). More and diverse latent GPs enhance the extraction of multi-scale features shared across related tasks at the cost of however linearly increased time complexity as well as more hyperparameters to be inferred. Can we have the desired number of latent GPs (i.e., maintaining the acceptale model complexity) while enhancing the diversity of latent space, which is crucial for enabling high model expressivity to tackle complicated multi-task scenario?

To this end, we introduce an additional latent embedding that wraps the latent functions $\{f_q(.)\}_{q=1}^Q$ to express the model for the $c$-th output in a \textit{higher} $H$-dimensional space, which may induce informative statistical relations, as 
\begin{align} \label{eq_mogp_}
	y^c(\mbf{x}) = \sum_{h=1}^H a^c_h \bar{f}_h(\mbf{x}) + \epsilon^c,
\end{align}
where the additional latent embedding acts on $\{f_q(.)\}_{q=1}^Q$ as
\begin{align} \label{eq_bar_f_h}
\bar{f}_h(\mbf{x}) = \sum_{q=1}^Q b_q^h f_q(\mbf{x})
\end{align}
through linearly weighted combination. It is found that the new $\bar{f}_h(.)$ mixes the base GPs $\{f_q(.)\}_{q=1}^Q$, which could be regarded as the ``coordinate components'' to express the related outputs $\{y^c\}_{c=1}^C$, in a higher dimensional space ($H > Q$) before passing it to the following Gaussian likelihood. Now we obtain more latent functions $\{\bar{f}_h\}_{h=1}^H$ that are expected to induce more powerful feature extraction across tasks. 

But it is found through~\eqref{eq_bar_f_h} that these new latent functions are similar to each other due to the linear combination. For example, taking the extreme case with $Q=1$, we have almost the same functions $\{\bar{f}_h\}_{h=1}^H$, since each of them is a scaled version of the base GP $f_1$. This raises the degenerated modeling of related tasks in such an \textit{isotropic} latent space. Inspired from the idea of neural network (NN), Jankowiak and Gardner~\cite{jankowiak2019neural} proposed to additionally apply an activation function, for example, the ReLU activation, to the latent embeddings $\{\bar{f}_h\}_{h=1}^H$, see equation~\eqref{eq_nmogp}, which increases the non-linearity of latent functions but cannot alleviate the isotropic behavior of this high dimensional latent space.

Hence, to alleviate the above issue, the key is increasing the \textit{diversity} of $\{\bar{f}_h\}_{h=1}^H$. To this end, as depicted by Fig.~\ref{fig_LMCvsNLMC}(b), we propose to use the more flexible and powerful neural embedding of $\{f_q\}_{q=1}^Q$, i.e., making the mixing coefficient $b_q^h$ be dependent of input $\mbf{x}$, thus varying over the entire input domain and resulting in various latent functions $\{\bar{f}_h\}_{h=1}^H$ even with $Q=1$, the example of which has been illustrated in Fig.~\ref{fig_toy_Neural_study} on a toy case. It is worth noting that though the model capability of LMC could be improved in the high dimensional latent space, it still relies on the number and quality of base GPs.

So far, this improved LMC model can be written in the matrix form as
\begin{align} \label{eq_nsvlmc}
	\mbf{y}(\mbf{x}) = \mbf{A} \bar{\mbf{f}}(\mbf{x}) + \bm{\epsilon} = \mbf{A} \mbf{B}(\mbf{x}) \mbf{f}(\mbf{x}) + \bm{\epsilon},
\end{align}
where the likelihood mixture $\mbf{A} \in \mathbb{R}^{C \times H}$ performs task-specific regression, and the latent neural mixture $\mbf{B}(\mbf{x}) \in \mathbb{R}^{H \times Q}$ transforms the GPs into high-dimensional and diverse latent space. Given the model definition, the scalable variational inference of the proposed model will be elaborated in next section.

\section{Scalable variational inference} \label{sec_vi}
This section attempts to address the scalability of the proposed model and improve the quality of inference through advanced variational inference, followed by the discussions regarding the variants of neural embeddings for LMC.

\subsection{Tighter evidence lower bound}
To improve the scalability of the proposed MTGP model when handling massive data over tasks, we take the idea from sparse approximation by introducing a set of inducing variables $\mbf{u}_q \in \mathbb{R}^{M_q}$ as sufficient statistics for the latent function values $\mbf{f}_q \in \mathbb{R}^{N}$ at the pseudo inputs $\mbf{Z}_q \in \mathbb{R}^{M_q \times D}$ with $M_q \ll N$. We then let the inducing set $\mbf{u} = \{\mbf{u}_q\}_{q=1}^Q$ be a collection of the inducing variables for $Q$ independent latent functions.

For model inference, we introduce the following joint variational distribution 
\begin{align} \label{eq_q(f,u,A,B)}
q(\mbf{f}, \mbf{u}, \mbf{A}, \mbf{B}) = q(\mbf{A}) q(\mbf{B}) \prod_{q=1}^Q p(\mbf{f}_q|\mbf{u}_q) q(\mbf{u}_q)
\end{align}
to approximate the unknown, exact posterior $p(\mbf{f}, \mbf{u}, \mbf{A}, \mbf{B} | \mbf{y})$. In~\eqref{eq_q(f,u,A,B)}, the variational posterior $q(\mbf{u}_q) = \mathcal{N}(\mbf{u}_q | \mbf{m}_q, \mbf{S}_q)$ takes the Gaussian form with the mean $\mbf{m}_q$ and variance $\mbf{S}_q$ as hyperparameters to be inferred from data. Besides, under the multi-variate Gaussian prior assumption, we have the following Gaussian posterior
\begin{align}
p(\mbf{f}_q|\mbf{u}_q) = \mathcal{N}(\mbf{f}_q | \mbf{K}_{XZ_q} \mbf{K}^{-1}_{Z_q} \mbf{u}_q, \mbf{K}_q - \mbf{K}_{XZ_q} \mbf{K}^{-1}_{Z_q} \mbf{K}_{XZ_q}^{\mathsf{T}}), 
\end{align}
where the covariances $\mbf{K}_{Z_q} = k_q(\mbf{Z}_q, \mbf{Z}_q) \in \mathbb{R}^{M_q \times M_q}$ and $\mbf{K}_{XZ_q} = k_q(\mbf{X}, \mbf{Z}_q) \in \mathbb{R}^{N \times M_q}$. We can further obtain the variational posteriors for $\mbf{f}_q$ by integrating $\mbf{u}_q$ out as
\begin{align}
q(\mbf{f}_q) &= \mathbb{E}_{q(\mbf{u}_q)} [p(\mbf{f}_q|\mbf{u}_q)] = \mathcal{N}(\mbf{f}_q | \bm{\mu}_{q}, \bm{\Sigma}_{q}),
\end{align}
where the mean and covariance write respectively as
\begin{align}
\bm{\mu}_{q} &= \mbf{K}_{XZ_q} \mbf{K}^{-1}_{Z_q} \mbf{m}_q, \\
\bm{\Sigma}_{q} &= \mbf{K}_q + \mbf{K}_{XZ_q} \mbf{K}^{-1}_{Z_q} [\mbf{S}_q \mbf{K}^{-1}_{Z_q} - \mbf{I}] \mbf{K}_{XZ_q}^{\mathsf{T}}.
\end{align}
Note that for the high-dimensional variational posteriors $\{q(\bar{\mbf{f}}_h)\}_{h=1}^H$, they are still Gaussians because of the linear mixture of Gaussian posteriors $\{q(\mbf{f}_q)\}_{q=1}^Q$ according to~\eqref{eq_bar_f_h}.

Thereafter, we minimize the KL divergence $\mathrm{KL}[q(\mbf{f}, \mbf{u}, \mbf{A}, \mbf{B})||p(\mbf{f}, \mbf{u}, \mbf{A}, \mbf{B}|\mbf{y})]$, which is equivalent to maximize the evidence lower bound (ELBO) expressed as
\begin{align} \label{eq_elbo_L}
	\begin{aligned}
		\mathcal{L} =& \mathbb{E}_{q(\mbf{f}) q(\mbf{A}) q(\mbf{B})} [\log p(\mbf{y}|\mbf{f}, \mbf{A}, \mbf{B})] \\
		&-\mathrm{KL}[q(\mbf{u})||p(\mbf{u})] -\mathrm{KL}[q(\mbf{A})||p(\mbf{A})] -\mathrm{KL}[q(\mbf{B})||p(\mbf{B})].
	\end{aligned}
\end{align}
Note that the latent mixture $\mbf{B}$ here is input-dependent. Furthermore, we could derive a \textit{tighter} ELBO to improve the inference quality. To this end, we first obtain the lower bound for the log marginal likelihood $\log p(\mbf{y}|\mbf{A},\mbf{B})$ conditioned on the mixtures $\mbf{A}$ and $\mbf{B}$ as
\begin{align}
	\begin{aligned}
		\log p(\mbf{y}|\mbf{A},\mbf{B}) \ge \mathcal{L}_{\mbf{AB}} =& \mathbb{E}_{q(\mbf{f})} [\log p(\mbf{y}|\mbf{f}, \mbf{A}, \mbf{B})] -\mathrm{KL}[q(\mbf{u})||p(\mbf{u})] \\
		=& \tilde{\mathcal{L}}_{\mbf{AB}} -\mathrm{KL}[q(\mbf{u})||p(\mbf{u})].
	\end{aligned}
\end{align}
Thereafter, we arrive at the tighter bound for $\log p(\mbf{y})$ as
\begin{align}
	\begin{aligned} \label{eq_elbo_tight}
		\log p(\mbf{y}) \ge& \log \mathbb{E}_{p(\mbf{A}) p(\mbf{B})} [\exp(\mathcal{L}_{\mbf{AB}}) ] \\
		=& \log \mathbb{E}_{q(\mbf{A}) p(\mbf{B})} \left[\exp(\mathcal{L}_{\mbf{AB}}) \frac{p(\mbf{A})}{q(\mbf{A})} \right] \\
		\ge& \mathbb{E}_{q(\mbf{A})} \left[ \log \mathbb{E}_{p(\mbf{B})}\left[ \exp(\tilde{\mathcal{L}}_{\mbf{AB}}) \right] \right] - \mathrm{KL}[q(\mbf{A}) || p(\mbf{A})] \\
		&- \mathrm{KL}[q(\mbf{u}) || p(\mbf{u})]\\
		=& \mathcal{L}_{\mathrm{tight}} \ge \mathcal{L},
	\end{aligned}
\end{align}
which supports efficient stochastic variational inference (SVI) since the first expectation could be factorized over data points, see the detailed expressions for the components in ELBO~\eqref{eq_elbo_tight} in Appendix~\ref{app_elbo}. Hence, an unbiased estimation of $\mathcal{L}_{\mathrm{tight}}$ in the efficient mini-batch fashion can be obtained on a subset $\mathcal{B}^c$ of the $c$-th training data $\mbf{X}^c$ with $|\mathcal{B}^c| < N^c$. It is found that this ELBO achieves tighter lower bound by directly using the prior $p(\mbf{B})$ rather than the variational posterior $q(\mbf{B})$ in~\eqref{eq_elbo_L}. Besides, the usage of $q(\mbf{B})$ in $\mathcal{L}$ introduces additional hyperparameters as well as more inference efforts.

Particularly, the ELBO~\eqref{eq_elbo_tight} employs the \textit{factorized} Gaussian prior for the latent neural mixture
\begin{align} \label{eq_p_B}
	p(\mbf{B}) = \mathcal{N}(\mathrm{vec}(\mbf{B})|\bm{\mu}_{\mbf{B}}, \mathrm{diag}(\bm{\nu}_{\mbf{B}})).
\end{align}
Furthermore, to devise an informative prior for $\mbf{B}$ that is assumed to be dependent of input $\mbf{x}$, we could adopt the so-called \textit{neural embedding}, i.e., making the mean $\bm{\mu}_{\mbf{B}}$ and variance $\bm{\nu}_{\mbf{B}}$ in $p(\mbf{B})$ be parameterized as the function of input through multi-layer perceptron (MLP) as
\begin{align}
	[\bm{\mu}_{\mbf{B}}, \, \bm{\nu}_{\mbf{B}}] = \mathrm{MLP}_{\bm{\theta}}(\mbf{x}),
\end{align}
where $\bm{\theta}$ is the hyperparameters of the adopted MLP to be inferred from data. Finally, as for the likelihood mixing coefficients $\mbf{A}$, we have the following Gaussian prior and posterior
\begin{align}
	p(\mbf{A}) = \mathcal{N}(\mathrm{vec}(\mbf{A})|\mbf{0}, \mbf{I}), \quad q(\mbf{A}) = \mathcal{N}(\mathrm{vec}(\mbf{A})|\bm{\mu}_{\mbf{A}}, \mathrm{diag}(\bm{\nu}_{\mbf{A}})),
\end{align}
wherein the mean $\bm{\mu}_{\mbf{A}}$ and variance $\bm{\nu}_{\mbf{A}}$ will be freely inferred from data.

Moreover, we could further improve the quality of ELBO~\eqref{eq_elbo_tight} by using importance-weighted variational inference (IWVI)~\cite{burda2016importance, domke2018importance}, which attempts to decrease the variance by letting the term inside the expectation concentrated around its mean. To this end, the $\mbf{B}$ term inside the expectation is replaced with a sample average of $S$ terms $\{\mbf{B}^s\}_{s=1}^S$ as
\begin{align} \label{eq_elbo_iwvi}
\begin{aligned}
\mathcal{L}_{\mathrm{IWVI}} =& \mathbb{E}_{q(\mbf{A})} \left[ \log \frac{1}{S} \sum_{s=1}^S \mathbb{E}_{p(\mbf{B}^{s})}\left[ \exp(\tilde{\mathcal{L}}_{\mbf{AB}^s}) \right] \right] \\
& - \mathrm{KL}[q(\mbf{A}) || p(\mbf{A})] - \mathrm{KL}[q(\mbf{u}) || p(\mbf{u})].
\end{aligned}
\end{align}
This has been found to be a strictly tighter bound than $\mathcal{L}_{\mathrm{tight}}$ in~\eqref{eq_elbo_tight}~\cite{burda2016importance,cremer2017reinterpreting}. This tight and compacted ELBO can be estimated through the reparameterization trick introduced in~\cite{kingma2014auto}.

So far, we have provided the scalable and effective variational inference strategy for the proposed MTGP model. This improved stochastic variational LMC model using neural embedding of coregionalization is donated as NSVLMC.

\subsection{Predictions}
Given the inferred hyperparameters of proposed NSVLMC, we then perform multi-task prediction at unseen points. First, the predictions of $Q$ latent independent GPs at the test point $\mbf{x}_*$ are
\begin{align}
	q(\mbf{f}_{*}) = \int \prod_{q=1}^Q p(f_{q*}|\mbf{u}_q) q(\mbf{u}_q) d\mbf{u}_q = \mathcal{N}(\mbf{f}_{*}| \bm{\mu}_{\mbf{f}*}, \bm{\nu}_{\mbf{f}*}),
\end{align}
where the mean and variance are respectively expressed as
\begin{align}
	[\bm{\mu}_{\mbf{f}*}]_q &= k_q(\mbf{x}_*,\mbf{Z}_q) \mbf{K}^{-1}_{Z_q} \mbf{m}_q, \\
	[\bm{\nu}_{\mbf{f}*}]_q &= k_q(\mbf{x}_*,\mbf{x}_*) + k_q(\mbf{x}_*,\mbf{Z}_q) \mbf{K}^{-1}_{Z_q} [\mbf{S}_q \mbf{K}^{-1}_{Z_q} - \mbf{I}] k_q^{\mathsf{T}}(\mbf{x}_*,\mbf{Z}_q),
\end{align}
for $1 \le q \le Q$. Thereafter, the conditional predictions for the observations are
\begin{align}
	q(\mbf{y}_*|\mbf{A}, \mbf{B}_*) = \int p(\mbf{y}_*|\mbf{f}_{*}) q(\mbf{f}_{*}) d\mbf{f}_* = \mathcal{N}(\mbf{y}_*|\bm{\mu}_*, \bm{\Sigma}_*),
\end{align}
where
\begin{align}
	\bm{\mu}_* &= \mbf{A} \mbf{B}_* \bm{\mu}_{\mbf{f}*}, \\
	\bm{\Sigma}_* &= \mbf{A} \mbf{B}_* \bm{\Sigma}_{\mbf{f}*} \mbf{B}_*^{\mathsf{T}} \mbf{A}^{\mathsf{T}} + \bm{\Sigma}_{\epsilon},
\end{align}
with the diagonal covariance $\bm{\Sigma}_{\mbf{f}*} = \mathrm{diag}[\bm{\nu}_{\mbf{f}*}]$ and the task-related noise variances $\bm{\Sigma}_{\epsilon} = \mathrm{diag}[\nu_{\epsilon}^1, \cdots, \nu_{\epsilon}^C]$.

Finally, we integrate the randoms $\mbf{A}$ and $\mbf{B}_*$ out to derive the final predictions
\begin{align} \label{eq_y*}
p(\mbf{y}_*|\mbf{y}) \approx \int q(\mbf{y}_*|\mbf{A}, \mbf{B}_*) q(\mbf{A}) p(\mbf{B}_*) d\mbf{A}d\mbf{B}_*,
\end{align}
which can be estimated through Markov Chain Monte Carlo (MCMC) sampling. It is worth noting that the predictive distribution $p(\mbf{y}_*|\mbf{y})$ is no longer Gaussian.

\subsection{Discussions} \label{sec_diss}
\begin{figure}[t!]
	\centering
	\includegraphics[width=0.8\textwidth]{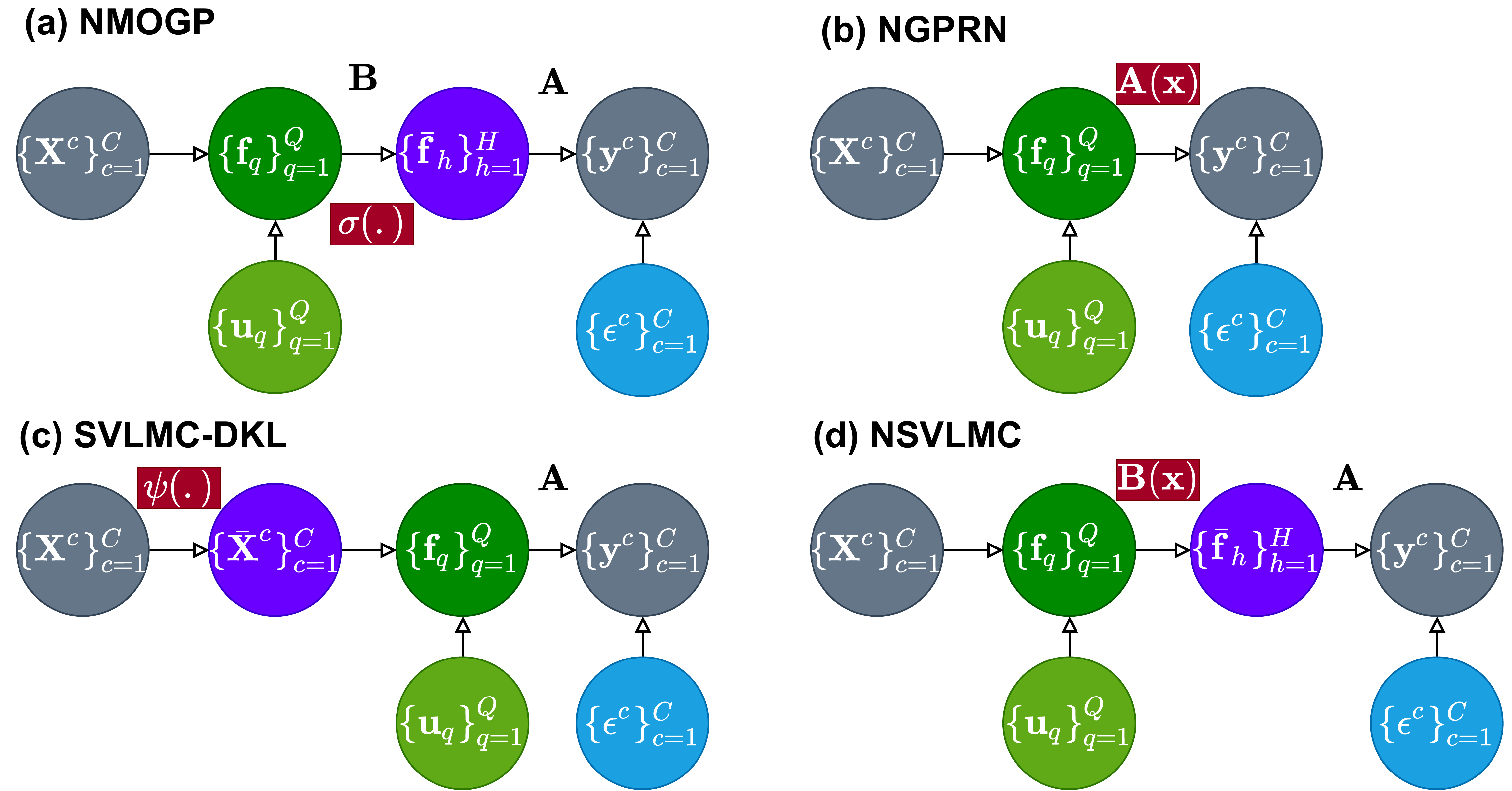}
	\caption{The graphical models of various latent neural embeddings for scalable LMCs including (a) NMOGP, (b) NGPRN, (c) SVLMC-DKL, and (d) the proposed NSVLMC. Note that these neural embeddings are performed through the terms marked in red background.}
	\label{fig_neuralEmbeddings}
\end{figure}

Except for the proposed neural embedding acting on the latent GPs $\{f_q\}_{q=1}^Q$ in~\eqref{eq_nsvlmc}, how about other neural embeddings? For example, inspired by the idea from Gaussian process regression network (GPRN)~\cite{wilson2012gaussian}, we could perform neural embedding on the weights in likelihood mixture $\mbf{A}$, thus arriving at the so-called neural GPRN (NGPRN)~\cite{jankowiak2019neural} as
\begin{align} \label{eq_ngprn}
	\mbf{y}(\mbf{x}) = \mbf{A}(\mbf{x}) \mbf{f}(\mbf{x}) + \bm{\epsilon},
\end{align}
where the neural embedding $\mbf{A}(\mbf{x}) = \mathrm{MLP}_{\bm{\alpha}}(\mbf{x})$ with the NN parameters $\bm{\alpha}$ to be inferred. It is observed that in comparison to the conventional input-independent likelihood mixture $\mbf{A}$, now the new $\mbf{A}(\mbf{x})$ controlled by a neural network varies over the input domain in order to express the flexible point-by-point similarity across tasks, thus greatly enhancing the ability for tackling complicated multi-task regression. The introduction of neural network assisted similarity in the task-specific level however may raise the issue of poor generalization, especially when some of the tasks have a few number of data points, see the numerical experiments in section~\ref{sec_exp}.

Alternatively, observing that the transformation of $\mbf{f}(\mbf{x})$ in~\eqref{eq_nsvlmc} is similar to the architecture of neural network, we could apply a nonlinear activation function $\sigma(.)$, e.g., the tanh function, to the linear mapping $\mbf{B} \mbf{f}(\mbf{x})$ instead of making $\mbf{B}$ be dependent of input $\mbf{x}$ as~\cite{jankowiak2019neural}
\begin{align} \label{eq_nmogp}
	\mbf{y}(\mbf{x}) = \mbf{A} \sigma(\mbf{B} \mbf{f}(\mbf{x})) + \bm{\epsilon}.
\end{align}
As has been discussed before, this model, denoted as NMOGP, however does not improve the diversity of the $H$-dimensional latent space, which is crucial for enhancing model capability. Besides, the nonlinear activation makes $\sigma(\mbf{B} \mbf{f}(\mbf{x}))$ no longer Gaussians.

Finally, inspired by the idea of deep kernel learning (DKL)~\cite{wilson2016deep, liu2021deep}, we could perform embedding on the inputs $\mbf{X}$ before passing them to the following LMC model as
\begin{align}
	\mbf{y}(\mbf{x}) = \mbf{A} \mbf{f}(\mathrm{MLP}_{\bm{\psi}}(\mbf{x}))+ \bm{\epsilon},
\end{align}
where the neural network wrapper $\mathrm{MLP}_{\bm{\psi}}(\mbf{x})$ encodes the input feature into a latent space, which is expected to improve the subsequent LMC modeling. This model using sparse approximation is denoted as SVLMC-DKL. The benefits brought by additional input transformation however is insignificant in the following numerical experiments.

The superiority, for example, higher quality of predictions and better generalization, of our proposed neural embedding against the above neural embedding variants will be verified empirically through extensive numerical experiments in section~\ref{sec_exp}.

\section{Numerical experiments} \label{sec_exp}
This section first investigates the methodological characteristics of the proposed NSVLMC on a three-task toy case, followed by the comprehensive comparison study against existing LMCs on three small- and large-scale real-world multi-task datasets. Finally, it specifically applies the proposed NSVLMC to the cross-fluid modeling of unsteady fluidized bed.

\subsection{Toy case}
We first study the methodological characteristics of the proposed NSVLMC on a toy case.  This toy case has three tasks generated from the same four latent functions expressed respectively as
\begin{align}
\begin{aligned}
y_1(x) =&  0.5f_1(x) - 0.4 f_2(x) + 0.6f_3(x) + 0.6f_4(x) + \epsilon, \\
y_2(x) =& -0.3f_1(x) + 0.43f_2(x) - 0.5f_3(x) + 0.1f_4(x) + \epsilon, \\ 
y_3(x) =&  1.5f_1(x)              + 0.3f_3(x) + 0.6f_4(x) + \epsilon,  
\end{aligned}
\end{align}
where the independent noise $\epsilon \sim \mathcal{N}(\epsilon|0, 0.04)$, and the four latent functions are
\begin{align}
\begin{aligned}
f_1(x) =& 0.5\sin (3x) + x, \\
f_2(x) =& 3\cos(x) - x, \\ 
f_3(x) =& 2.5\cos(5x-1), \\
f_4(x) =& \sin(1.5x).
\end{aligned}
\end{align}
We randomly generate 100 points in the one-dimensional range $[-5, 5]$ for outputs $y_1$ and $y_3$, and 10 random points for $y_2$. Through modeling the three related tasks simultaneously, the knowledge transfer across tasks is expected to be extracted for improving the modeling of $y_2$ with a few number of data points.

We first investigate the modeling quality of the proposed NSVLMC with different numbers of latent functions, i.e., using different $Q$ values, on this toy case. We employ (i) the SE kernel and $M_q=25$ inducing variables for each latent function $f_q$; (ii) $H=100$ hidden latent functions $\{\bar{f}_h\}_{h=1}^H$; and (iii) the Adam optimizer to train the model over 20000 iterations. Other detailed model configurations are provided in Appendix~\ref{sec_exp_details}. The original SVLMC model has also been introduced as baseline for comparison. 

\begin{figure}[t!]
	\centering
	\includegraphics[width=1.0\textwidth]{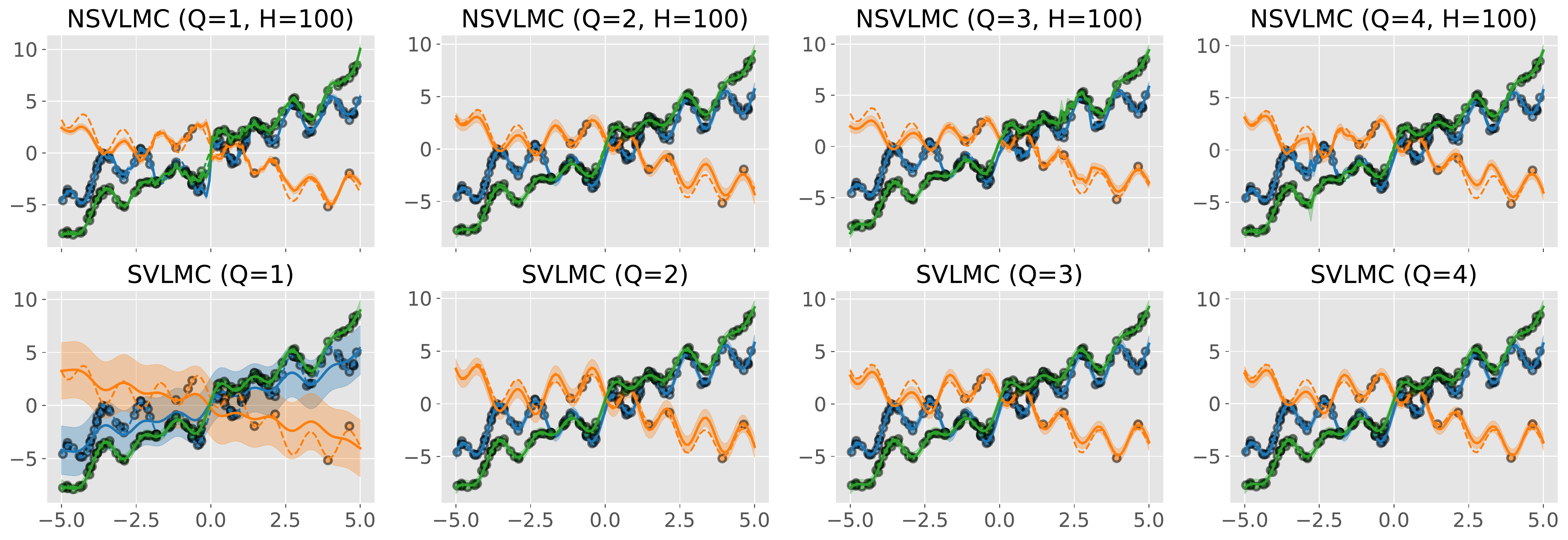}
	\caption{The multi-task regression of NSVLMC and SVLMC with increasing $Q$ on the toy case. The dashed curves marked in different colors represent the latent true responses of three tasks. The curves together with the shaded region represent 95\% predictive distributions. The dots marked in different colors are the training data for three tasks.}
	\label{fig_toy_NSVLMC}
\end{figure}

\begin{figure}[t!]
	\centering
	\includegraphics[width=1.0\textwidth]{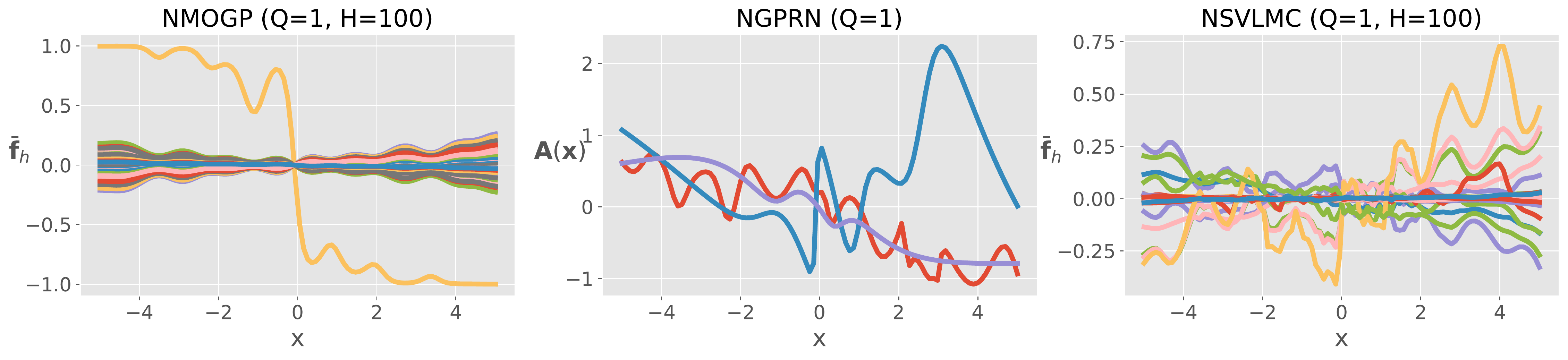}
	\caption{The latent functions or weights obtained by neural embeddings in NMOGP, NGPRN and NSVLMC on the toy case.}
	\label{fig_toy_Neural_study}
\end{figure}

Fig.~\ref{fig_toy_NSVLMC} depicts the comparative results of NSVLMC and SVLMC with increasing $Q$ on the toy case. It is interesting to observe that when $Q=1$, i.e., we are only using a single latent function, the original SVLMC of course cannot model the three tasks well since the outputs are limited to the scaled version of the single latent function, leaving no flexibility to explain the task-related features. Contrarily, the proposed NSVLMC first transforms the single latent function into a high-dimensional space to have $H=100$ hidden latent functions with diverse characteristics (see Fig.~\ref{fig_toy_Neural_study}), which thereby help well model the three tasks even with $Q=1$. With the increase of $Q$, the diverse base latent functions quickly improve the model capability of both SVLMC and NSVLMC. For example, by transferring shared knowledge from $y_1$ and $y_3$, the SVLMC equipped with $Q=2$ latent functions well models $y_2$ even in the left input domain with unseen data points.

\begin{figure}[t!]
	\centering
	\includegraphics[width=1.0\textwidth]{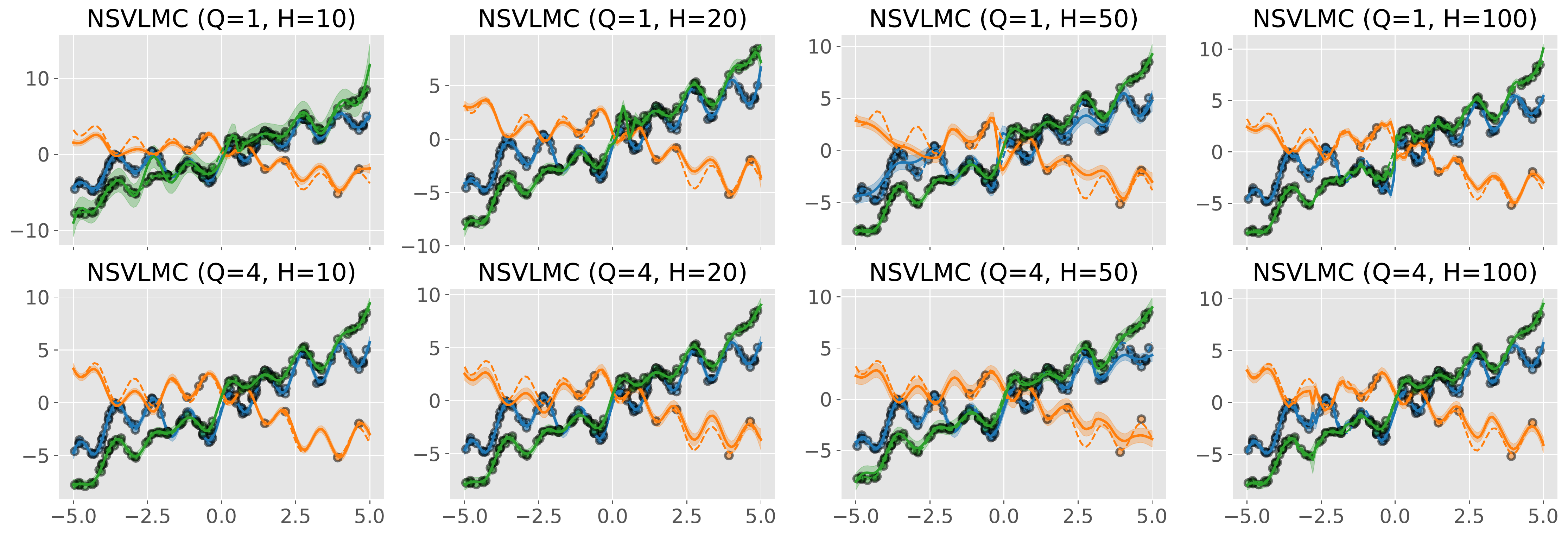}
	\caption{The impact of $H$ on the multi-task regression of NSVLMC on the toy case. The dashed curves marked in different colors represent the latent true responses of three tasks. The curves together with the shaded region represent 95\% predictive distributions. The dots marked in different colors are the training data for three tasks.}
	\label{fig_toy_NSVLMC_H}
\end{figure}

The performance of NSVLMC with $Q=1$ and $H=100$ is impressive on the toy case. Hence, we thereafter investigate the impact of $H$ on the performance of NSVLMC in Fig.~\ref{fig_toy_NSVLMC_H}. As for the challenging case with $Q=1$, it is found that the increase of $H$ generally improves the model capability since it describes the multi-scale task features more finely. As for the case $Q=4$, due to the sufficient and diverse latent base GPs, the increase of $H$ does not bring significant improvement on this toy case. Particularly, it is found that the enhanced model expressivity with $H=100$ have raised slight over-fitting on the toy case with $Q=4$. This implies that the setting of $H$ is related to $Q$: when $Q$ is small, indicating a poor and limited base GP set, we need a large $H$ to have diverse hidden latent functions for improving the prediction; contrarily, when $Q$ is large, indicating sufficient and diverse base GPs, we need a mild $H$ to preserve the model capability while alleviating the issue of over-fitting.

\begin{figure}[t!]
	\centering
	\includegraphics[width=1.0\textwidth]{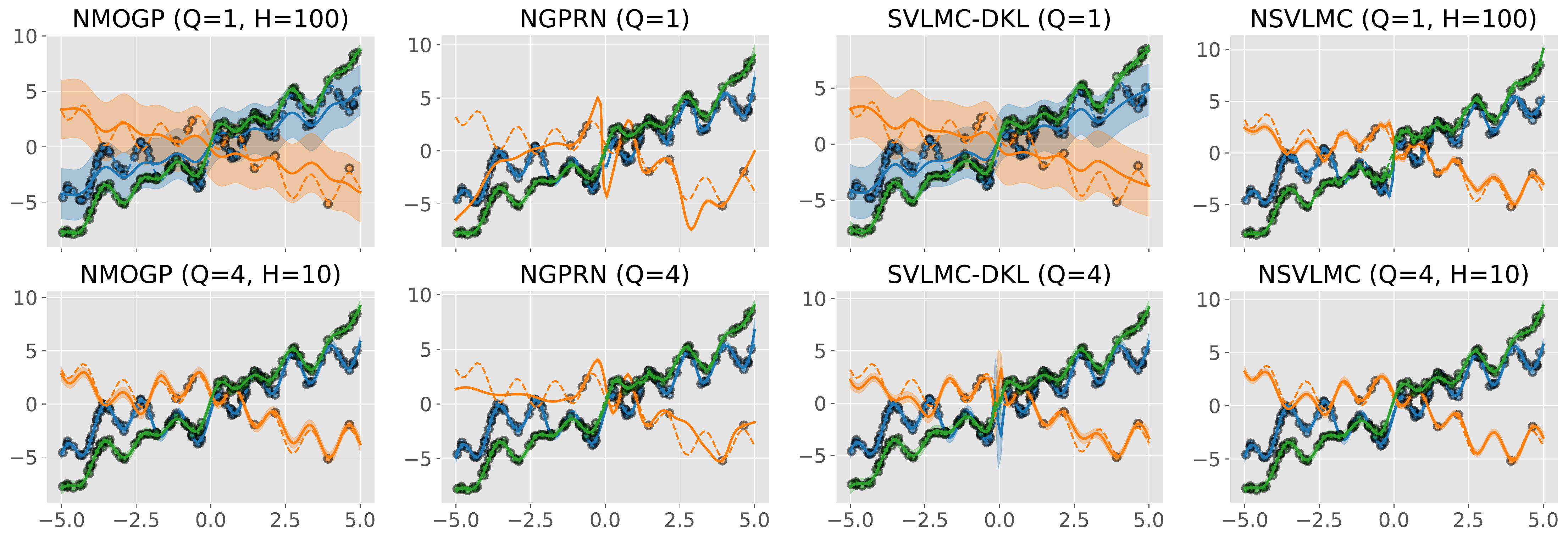}
	\caption{Comparison of different neural embeddings on the toy case. The dashed curves marked in different colors represent the latent true responses of three tasks. The curves together with the shaded region represent 95\% predictive distributions. The dots marked in different colors are the training data for three tasks.}
	\label{fig_toy_Neural_Embedding}
\end{figure}

Finally, as has been discussed in section~\ref{sec_diss}, variants of neural embeddings could be implemented for LMC. Therefore, we here compare them against the proposed NSVLMC on the toy case in Fig.~\ref{fig_toy_Neural_Embedding}. As for NMOGP, it adopts global weights $\mbf{A}$ and $\mbf{B}$ but applies nonlinear activation to the transform $\mbf{B}\mbf{f}(\mbf{x})$, which however does not well improve the diversity of $\bar{\mbf{f}}(\mbf{x})$ in the $H$-dimensional space, as shown in Fig.~\ref{fig_toy_Neural_study}. Therefore, the NMOGP fails to perform multi-task regression when $Q=1$. As for SVLMC-DKL, it performs an input transformation through neural networks before multi-task modeling. This has no impact on the diversity of latent functions, thus resulting in poor performance when $Q=1$. Besides, the free neural input encoding may raise the issue of over-fitting, see the model with $Q=4$ and the discussions in~\cite{liu2021deep}. As for NGPRN, instead of improving the diversity of latent functions, it attempts to employ task-specific, input-dependent weights $\mbf{A}(\mbf{x})$ to tackle challenging multi-task cases. Consequently, we can find in~Fig.~\ref{fig_toy_Neural_Embedding} that the NGPRN well fits all the outputs at training points when $Q=1$, but raises obvious over-fitting for $y_2$ and poor generalization in the left input domain. This is due to the flexible and task-specific weights $\mbf{A}(\mbf{x})$ parameterized by NN. As shown in Fig.~\ref{fig_toy_Neural_study}, the powerful neural network is easy to become over-fitting when we have scarce training data. Finally, for the proposed NSVLMC, it performs well in the studied scenarios due to the flexible and diverse hidden latent functions $\bar{\mbf{f}}(\mbf{x})$ in comparison to that of NMOGP, see Fig.~\ref{fig_toy_Neural_study}. Besides, though equipped with neural networks for the latent neural mixture $\mbf{B}(\mbf{x})$, the NSVLMC does not obviously suffer from the issue of over-fitting. This may be attributed to two reasons: (i) the neural embedding is conducted on the task-sharing level, i.e., it is trained on the inputs of all the tasks, see Fig.~\ref{fig_LMCvsNLMC}(b); and (ii) the model is implemented in the complete Bayesian framework, which is beneficial for guarding against over-fitting.

\subsection{Real-world datasets}
This section verifies the superiority of the proposed NSVLMC on three multi-scale multi-task datasets, which will be elaborated as below. The detailed experimental configurations are provided in Appendix~\ref{sec_exp_details}. Besides, the error criteria adopted in the following comparison for model evaluation are provided in Appendix~\ref{sec_err_criteria}.

\subsubsection{Datasets description}

\texttt{Jura} dataset.\footnote{The data is available at~\url{https://sites.google.com/site/goovaertspierre/pierregoovaertswebsite/download/}.
} This geospatial dataset describes the heavy metal concentration of Cadmium, Nickel and Zinc measured from the topsoil at 359 positions in a local region of the Swiss Jura. We follow the experimental settings in~\cite{requeima2019gaussian}: we observe the Nickel and Zinc outputs at all the $359$ positions and the Cadmium output at the first 259 positions, with the goal being to predict the Cadmium output at the remaining 100 positions. For the fair of comparison, we adopt the mean absolute error (MAE) and the negative log likelihood (NLL) criteria to quantify the model performance.

\texttt{EEG} dataset.\footnote{The data is available at~\url{https://archive.ics.uci.edu/ml/datasets/eeg+database}.
} This medical dataset describes the voltage readings of seven electrons FZ and F1-F6 placed on the patient's scalp over $t=1$s measurement period at 256 discrete time points. We follow the experimental settings in~\cite{requeima2019gaussian}: we observe the whole signals of F3-F6 and the first 156 signals of FZ, F1 and F2, with the goal being to predict the last 100 signals of FZ, F1 and F2. For the fair of comparison, we adopt the standardized mean square error (SMSE) and the NLL criteria to quantify the model performance.

\texttt{Sarcos} dataset.\footnote{The data is available at~\url{http://www.gaussianprocess.org/gpml/data/}.} This large-scale, high dimensional data is related to the inverse dynamic modeling of a 7-degree-of-freedom anthropomorphic robot arm~\cite{vijayakumar2000locally} with 21 input variables (7 joints positions, 7 joint velocities, and 7 joint accelerations) and the corresponding 7 joint torques as outputs. We build three cases with different scales from this dataset for investigating multi-task modeling. The first two cases (cases A and B) study the joint modeling of the 4th torque and the 7th torque, the spearman correlation of which is up to $r=0.96$, indicating the highest similarity between two outputs. We split the data so that the 7th torque has 44484 data points. But for case A, the 4th torque only have 50 data points; while for case B, the 4th torque has 2000 data point. The final case C attempts to simultaneously model the 6th and 7th torques, which have the lowest and negative spearman correlation as $r=-0.10$. Similarly, we observe 44484 data points for the 7th torque and 2000 data points for the 4th torque. For the above three cases, we have a separate test set consisting of 4449 data points for model evaluation. Similar to the \texttt{EEG} dataset, we employ the SMSE and NLL criteria for performance quantification.

It is found that the \texttt{Jura} dataset and the three cases of \texttt{Sarcos} perform multi-task \textit{interpolation}, while the \texttt{EEG} dataset performs the challenging multi-task \textit{extrapolation}. Besides, the target outputs of the \texttt{Jura} dataset, the \texttt{EEG} dataset and case A of the \texttt{Sarcos} dataset are small-scale with no more than 100 data points for target outputs; while the training size of target outputs of cases B and C of \texttt{Sarcos} is up to 2000, which is a challenging scenario for LMC to outperform single task GP.

\subsubsection{Results and discussions}
This comparative study introduces state-of-the-art LMCs as well as other MTGP competitors, including (i) the Gaussian processes autoregressive regression with nonlinear correlations (GPAR-NL)~\cite{requeima2019gaussian}, which has been verified to be superior in comparison to previous MTGPs, for example, CoKriging~\cite{wackernagel2003multivariate}, intrinstic coregionalisation model (ICM)~\cite{williams2007multi}, semiparametric latent factor model (SLFM)~\cite{teh2005semiparametric}, collaborative multi-output GP (CGP)~\cite{nguyen2014collaborative}, convolved multi-output GP (CMOGP)~\cite{alvarez2011computationally}, and GPRN~\cite{wilson2012gaussian}; (ii) the multi-output GPs with neural likelihoods~\cite{jankowiak2019neural}, including NMOGP, NGPRN and SVLMC-DKL; and (iii) the sparse GP variational autoencoders (GP-VAE)~\cite{ashman2020sparse} with partial inference networks~\cite{vedantam2018generative}. Besides, the baselines GP and stochastic variational GP (SVGP)~\cite{hensman2013gaussian} are involved in the comparison. Tables~\ref{tab_jura}-\ref{tab_sarcos} showcase the comparative results of different models on the three multi-task datasets, respectively. It is notable that the results of GP and GPAR-NL in Tables~\ref{tab_jura} and~\ref{tab_eeg} are taken from~\cite{requeima2019gaussian}, and the results of GP-VAE are taken from~\cite{ashman2020sparse}. We have the following findings from the comparative results.

\begin{table}
	\caption{Comparative results on the \texttt{Jura} dataset, with the best and second-best results marked in gray and light gray, respectively.} 
	\label{tab_jura}
	\centering
	\resizebox{0.8\columnwidth}{!}{%
	\begin{tabular}{lrrrr}
		\hline
		Models &GP &GPAR-NL &GP-VAE &SVLMC \\
		\hline
		MAE &0.5739 &0.4324 &\cellcolor{lightgray}0.40$_{\pm0.01}$ &0.4580$_{\pm0.0047}$ \\
		NLL &NA &NA &1.0$_{\pm0.06}$ &\cellcolor{mygray}0.9686$_{\pm0.0100}$ \\
		\hline
		\hline
		Models &NMOGP &NGPRN &SVLMC-DKL &NSVLMC \\
		\hline
		MAE &0.4618$_{\pm0.0036}$ &0.5619$_{\pm0.0187}$ &0.5173$_{\pm0.0136}$ &\cellcolor{mygray}0.4196$_{\pm0.0077}$ \\
		NLL &1.0172$_{\pm0.0090}$ &1.2887$_{\pm0.0859}$ &1.0335$_{\pm0.0208}$ &\cellcolor{lightgray}0.8681$_{\pm0.0263}$ \\
		\hline
	\end{tabular}
	}
\end{table}

\textbf{MTGPs outperform (SV)GP for correlated tasks with scarce data.} It is found that by leveraging the similarity among tasks, the MTGPs could transfer knowledge from other tasks in order to improve the modeling quality of target tasks even with scarce data, see for example the results on \texttt{Jura}, \texttt{EEG} and case A of \texttt{Sarcos} in Tables~\ref{tab_jura}-\ref{tab_sarcos}. But when the target task has sufficient data points, the benefits brought by multi-task learning is insignificant, especially for the lowly correlated tasks, see the results of cases B and C of \texttt{Sarcos} in Table~\ref{tab_sarcos}.

\begin{table}
	\caption{Comparative results on the \texttt{EEG} dataset, with the best and second-best results marked in gray and light gray, respectively.} 
	\label{tab_eeg}
	\centering
	\resizebox{0.8\columnwidth}{!}{%
		\begin{tabular}{lrrrr}
			\hline
			Models &GP &GPAR-NL &GP-VAE &SVLMC \\
			\hline
			SMSE &1.75 &0.26 &0.28$_{\pm0.04}$ &0.2335$_{\pm0.1183}$ \\
			NLL &2.60 &\cellcolor{lightgray}1.63 &2.23$_{\pm0.21}$ &1.7184$_{\pm0.3521}$ \\
			\hline
			\hline
            Models &NMOGP &NGPRN &SVLMC-DKL &NSVLMC \\
            \hline
            SMSE &\cellcolor{mygray}0.2148$_{\pm0.0554}$ &1.9012$_{\pm0.7016}$ &1.7883$_{\pm0.7998}$ &\cellcolor{lightgray}0.1783$_{\pm0.0185}$ \\
            NLL &1.8835$_{\pm0.3058}$ &8.9466$_{\pm4.6042}$ &3.0267$_{\pm0.6609}$ &\cellcolor{mygray}1.7035$_{\pm0.2409}$ \\
            \hline
		\end{tabular}
	}
\end{table}

\textbf{NGPRN risks over-fitting for scarce data and extrapolation.} It is observed that the NGPRN has the poorest performance on \texttt{Jura}, \texttt{EEG} and case A of \texttt{Sarcos}, especially in terms of the NLL criterion. The NGPRN introduces input-varying, task-related weights $\mbf{A}(\mbf{x})$ in~\eqref{eq_ngprn} to form the flexible and powerful neural likelihood, the capability of which has been illustrated in the toy case with $Q=1$. But this expressive likelihood mixture for tasks is easy to risk overfitting, thus resulting in poor prediction mean and overestimated variance, see for example the extremely large NLL for case A in Table~\ref{tab_sarcos}. The poor prediction of NGPRN becomes more serious for extrapolation, see the results on the \texttt{EEG} dataset and the illustration in Fig.~\ref{fig_toy_Neural_Embedding}. This is due to the poor generalization of task-specific neural network modeling of $\mbf{A}(\mbf{x})$ at unseen points, see the illustration in Fig.~\ref{fig_toy_Neural_study}. The above issues however could be alleviated by increasing the training size, see the desirable results of NGPRN for cases B and C of \texttt{Sarcos} in Table~\ref{tab_sarcos}. But as have been discussed before, the benefits of multi-task modeling in this large-scale scenario are insignificant.

\begin{table}
	\caption{Comparative results on the three cases of \texttt{Sarcos} dataset, with the best and second-best results marked in gray and light gray, respectively.} 
	\label{tab_sarcos}
	\centering
	\resizebox{\columnwidth}{!}{%
		\begin{tabular}{lrrrrrrr}
			\hline
			&Metric &SVGP &SVLMC &NMOGP &NGPRN &SVLMC-DKL &NSVLMC \\
			\hline
			\multirow{2}*{A} &SMSE &0.1397$_{\pm0.0143}$ &\cellcolor{mygray}0.0708$_{\pm0.0044}$ &0.0756$_{\pm0.0031}$ &0.1586$_{\pm0.0367}$ &0.0761$_{\pm0.0070}$ &\cellcolor{lightgray}0.0665$_{\pm0.0046}$ \\
			~ &NLL &2.8457$_{\pm0.0776}$ &\cellcolor{mygray}2.7476$_{\pm0.0506}$ &2.7568$_{\pm0.0210}$ &56.3878$_{\pm16.6132}$ &2.7637$_{\pm0.0526}$ &\cellcolor{lightgray}2.6902$_{\pm0.0442}$ \\
			\hline
			\hline
			\multirow{2}*{B} &SMSE &0.0235$_{\pm0.0009}$ &0.0278$_{\pm0.0112}$ &0.0284$_{\pm0.0147}$ &\cellcolor{lightgray}0.0139$_{\pm0.0018}$ &0.0221$_{\pm0.0026}$ &\cellcolor{mygray}0.0177$_{\pm0.0011}$ \\
			~ &NLL &2.3348$_{\pm0.0085}$ &2.2047$_{\pm0.1700}$ &2.2069$_{\pm0.1880}$ &\cellcolor{lightgray}1.8113$_{\pm0.0567}$ &2.1246$_{\pm0.0565}$ &\cellcolor{mygray}1.9807$_{\pm0.0311}$ \\
		    \hline
		    \hline
			\multirow{2}*{C} &SMSE &0.0951$_{\pm0.0025}$ &0.2445$_{\pm0.0743}$ &0.1628$_{\pm0.0433}$ &\cellcolor{mygray}0.0941$_{\pm0.0079}$ &0.1962$_{\pm0.0167}$ &\cellcolor{lightgray}0.0891$_{\pm0.0254}$ \\
			~ &NLL &\cellcolor{mygray}0.8238$_{\pm0.0128}$ &1.2129$_{\pm0.2073}$ &1.0309$_{\pm0.1701}$ &0.9536$_{\pm0.0732}$ &1.1408$_{\pm0.0395}$ &\cellcolor{lightgray}0.7119$_{\pm0.1530}$ \\
			\hline
		\end{tabular}
	}
\end{table}

\textbf{NSVLMC showcases superiority over counterparts.} In comparison to SVLMC, the SVLMC-DKL wraps inputs through neural networks, which may ease the subsequent latent GP regression but having limited enhancement for the learning of task similarities, thus resulting in slight improvements. Similar to the proposed NSVLMC, the NMOGP maps the $Q$ latent functions into a $H$-dimensional space and applies nonlinear activation before passing them to the likelihood. This helps it perform better than SVLMC on most cases. But this sort of LMC is still undesirable for tackling challenging scenarios due to the limited diversity of latent functions, see the illustration in Fig.~\ref{fig_toy_Neural_study}. As an improvement, the proposed NSVLMC adopts input-varying mixture $\mbf{B}(\mbf{x})$ to enhance the diversity of latent functions, thus showcasing the best results even for case C of the \texttt{Sarcos} dataset with the nearly uncorrelated tasks in Table~\ref{tab_sarcos}. It is worth noting that different from the input-varying, task-related $\mbf{A}(\mbf{x})$ in NGPRN, the mixture $\mbf{B}(\mbf{x})$ in NSVLMC is performed at the task-shared level, thus helping guarding against over-fitting.

\begin{figure*}[t!]
	\centering
	\begin{subfigure}
		\centering
		\includegraphics[width=.9\textwidth]{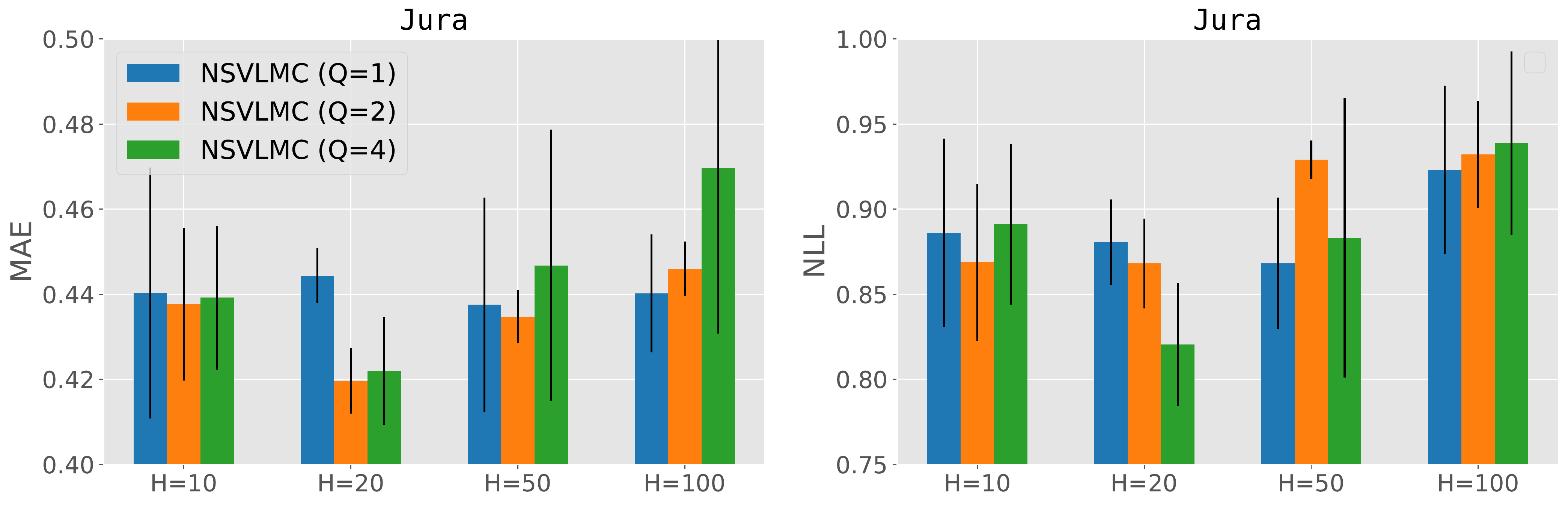}
	\end{subfigure}%
	\begin{subfigure}
		\centering
		\includegraphics[width=.9\textwidth]{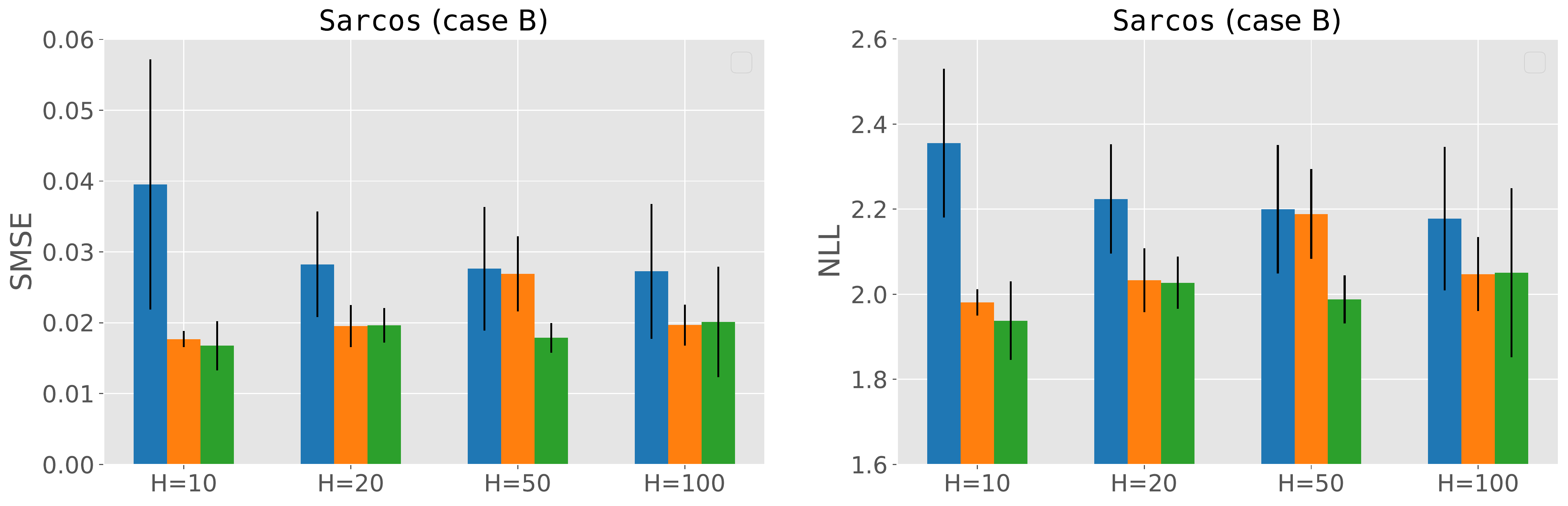}
	\end{subfigure}
	\caption{Impact of parameters $Q$ and $H$ on the performance of NSVLMC on the \texttt{Jura} dataset and case B of the \texttt{Sarcos} dataset.}
	\label{fig_impact_L_H} 
\end{figure*}

\textbf{The cooperation of parameters $Q$ and $H$ is crucial for NSVLMC.} We finally investigate the impact of parameters $Q$ and $H$ on the performance of NSVLMC on the \texttt{Jura} dataset and case B of the \texttt{Sarcos} dataset in Fig.~\ref{fig_impact_L_H}. As has been discussed in the toy case, increasing the parameters $Q$ and $H$ will enhance the expressivity of NSVLMC. Hence, for the small-scale \texttt{Jura} dataset, the performance of NSVLMC first increases with $Q$ and $H$ but thereafter becomes poor, which indicates the possibility of over-fitting due to the increasing model capability. This issue however has been alleviated on case B of the \texttt{Sarcos} dataset since the increased training data plays the role of regularization.

\subsection{Cross-fluid modeling for unsteady fluidized bed}
This section applies the proposed NSVLMC to an engineering case that is a gas-solid fluidized bed wherein the spherical glass beads are fluidized with air at ambient condition. The operating conditions and simulation inputs of this fluidized bed are consistent to the experiments conducted in~\cite{taghipour2005experimental}. The unsteady computational fluid dynamics (CFD) simulation is performed by using the twoPhaseEulerFoam solver of OpenFOAM~\cite{jasak2007openfoam} to get the volume fraction of particles (VFP) evolved over 20s time period within the bed.

We have two simulation cases, where case I is the fluidized bed filled up to $h=40$cm with glass beads, and case II is filled up to $h=38$cm. Since these two cases are only differ in the height of beads, we believe that the distributions of the spatial-temporal VFPs within the bed are similar. That is, we could adopt multi-task learning to predict the VFPs of these two cases. But directly modeling the high-dimensional VFP domain is unavailable for MTGP. Hence, we first resort to the well-known proper orthogonal decomposition (POD)~\cite{taira2017modal} to perform model reduction of VFP.\footnote{In machine learning community, it is known as principle component analysis (PCA).} Given $T=2000$ snapshots of the time-aware VFPs $\mbf{u}(\mbf{x}, t) \in \mathbb{R}^{N_m}$ of fluidized bed along time, the POD could extract the $R$-order dominant basis modes $\bm{\varphi}_k(\mbf{x}) \in \mathbb{R}^{N_m}$ as well as the time series coefficients $a_k(t)$ ($1 \le k \le R$) as
\begin{align} \label{eq_pod}
\mbf{u}(\mbf{x}, t) \approx \tilde{\mbf{u}}(\mbf{x}, t) = \sum_{k=1}^R a_k(t) \bm{\varphi}_k(\mbf{x}),
\end{align}
where $\tilde{\mbf{u}}(\mbf{x}, t)$ is the VFPs recovered by the $R$-order POD to approximate $\mbf{u}(\mbf{x}, t)$; $N_m = 6000$ (20 nodes $\times$ 300 nodes for the rectangular bed) is the number of structured meshes representing the discrete spatial approximation to the physics domain; the POD order satisfies $R \le \min\{T, N_m\}$ and the quality of VFP recovery increases with $R$; the input $\mbf{x}$ represents the input conditions of fluidized bed; $t$ ($0 \le t \le 20$) is an discrete time point within $[0, 20]$s; and finally, $a_k(t) = \int \mbf{u}(\mbf{x}, t) \bm{\varphi}_k(\mbf{x}) d\mbf{x}$ is the time series coefficient lying into a $R$-dimensional space.

Given the five-order ($R=5$) POD decomposition of the two fluidized bed cases,\footnote{We here only verify the superiority of LMCs over the single-task GP on the first five time series coefficients. It can be naturally extended for the modeling of remaining time series coefficients.} we obtain two coefficient sets $\{a_k^{\mathrm{I}}(t) \}_{1\le k\le R}^{0\le t\le 20}$ and $\{a_k^{\mathrm{II}}(t) \}_{1\le k\le R}^{0\le t\le 20}$. We setup the following multi-task experimental settings: for $\mbf{a}_k^{\mathrm{I}} = \{a_k^{\mathrm{I}}(t) \}_{t=0}^{20}$ ($1\le k \le R$), we observe the coefficients at all the 2000 time points, while for $\mbf{a}_k^{\mathrm{II}} = \{a_k^{\mathrm{II}}(t) \}_{t=0}^{20}$ ($1\le k \le R$), we only have observations at the first 20 time points, thus raising the demand of multi-task leaning to improve the predictions of case II at the remaining time points. Besides, we convert these time series coefficients into supervised leaning scenario through one-step ahead auto-regressive with the look-back window size of 10, thus resulting in 1990 data points for $\mbf{a}_k^{\mathrm{I}}$ and 10 data points for $\mbf{a}_k^{\mathrm{II}}$. Our goal is to predict the time series coefficients as well as recovering the VFPs $\tilde{\mbf{u}}^{\mathrm{II}}(\mbf{x}, t)$ using~\eqref{eq_pod} at the remaining 1980 time points for case II. It is worth noting that since we have five pairs of coefficients, five LMCs have been built. The detailed implementations are elaborated in Appendix~\ref{sec_exp_details}.

\begin{table}
	\caption{Comparative results on predicting the time series coefficients of case II of the fluidized bed, with the best and second-best results marked in gray and light gray, respectively.} 
	\label{tab_fluid_bed}
	\centering
	\resizebox{\columnwidth}{!}{%
		\begin{tabular}{lrrrrrrr}
			\hline
			&Metric &GP &SVLMC &NMOGP &NGPRN &SVLMC-DKL &NSVLMC \\
			\hline
			\multirow{2}*{$a_1^{\mathrm{II}}(t)$} &SMSE &0.1078 &0.0020$_{\pm0.0002}$ &\cellcolor{lightgray}0.0019$_{\pm0.0005}$ &0.0602$_{\pm0.0059}$ &0.0027$_{\pm0.0029}$ &\cellcolor{lightgray}0.0019$_{\pm0.0005}$ \\
			~ &NLL &-2.2103 &-5.4890$_{\pm0.0399}$ &\cellcolor{lightgray}-5.5215$_{\pm0.1186}$ &8.9083$_{\pm1.4626}$ &-5.3353$_{\pm0.7092}$ &\cellcolor{mygray}-5.5210$_{\pm0.1264}$ \\
			\hline
			\hline
			\multirow{2}*{$a_2^{\mathrm{II}}(t)$} &SMSE &1.0605 &\cellcolor{mygray}0.0007$_{\pm0.0001}$ &\cellcolor{lightgray}0.0006$_{\pm0.0001}$ &3.7819$_{\pm0.4869}$ &0.0009$_{\pm0.0001}$ &0.0007$_{\pm0.0003}$ \\
			~ &NLL &-2.1992 &\cellcolor{lightgray}-5.8037$_{\pm0.0332}$ &\cellcolor{mygray}-5.7904$_{\pm0.0294}$ &906.8614$_{\pm106.1105}$ &-5.6912$_{\pm0.0415}$ &-5.5145$_{\pm0.1767}$ \\
			\hline
			\hline
			\multirow{2}*{$a_3^{\mathrm{II}}(t)$} &SMSE &0.5182 &0.0014$_{\pm0.0004}$ &0.0021$_{\pm0.0004}$ &0.4628$_{\pm0.0433}$ &\cellcolor{mygray}0.0013$_{\pm0.0002}$ &\cellcolor{lightgray}0.0010$_{\pm0.0005}$ \\
			~ &NLL &-3.3703 &\cellcolor{mygray}-5.6414$_{\pm0.0942}$ &-5.4664$_{\pm0.0811}$ &108.4444$_{\pm10.4394}$ &-5.6404$_{\pm0.0658}$ &\cellcolor{lightgray}-5.7240$_{\pm0.1078}$ \\
			\hline
			\hline
\multirow{2}*{$a_4^{\mathrm{II}}(t)$} &SMSE &0.5535 &0.0017$_{\pm0.0005}$ &0.0018$_{\pm0.0009}$ &2.0475$_{\pm0.3549}$ &\cellcolor{mygray}0.0015$_{\pm0.0023}$ &\cellcolor{lightgray}0.0005$_{\pm0.0001}$ \\
~ &NLL &-3.0062 &-5.5750$_{\pm0.1148}$ &-5.5257$_{\pm0.1786}$ &486.9308$_{\pm085.3126}$ &\cellcolor{mygray}-5.4643$_{\pm0.2542}$ &\cellcolor{lightgray}-5.8115$_{\pm0.0150}$ \\
\hline
\hline
\multirow{2}*{$a_5^{\mathrm{II}}(t)$} &SMSE &0.7184 &0.0037$_{\pm0.0015}$ &0.0025$_{\pm0.0019}$ &0.3005$_{\pm0.3273}$ &\cellcolor{mygray}0.0021$_{\pm0.0009}$ &\cellcolor{lightgray}0.0015$_{\pm0.0024}$ \\
~ &NLL &-2.7060 &-5.1114$_{\pm0.3273}$ &-5.3924$_{\pm0.3760}$ &67.7025$_{\pm80.1554}$ &\cellcolor{mygray}-5.4629$_{\pm0.1816}$ &\cellcolor{lightgray}-5.6345$_{\pm0.5152}$ \\
\hline
		\end{tabular}
	}
\end{table}

\begin{figure}[t!]
	\centering
	\includegraphics[width=1.0\textwidth]{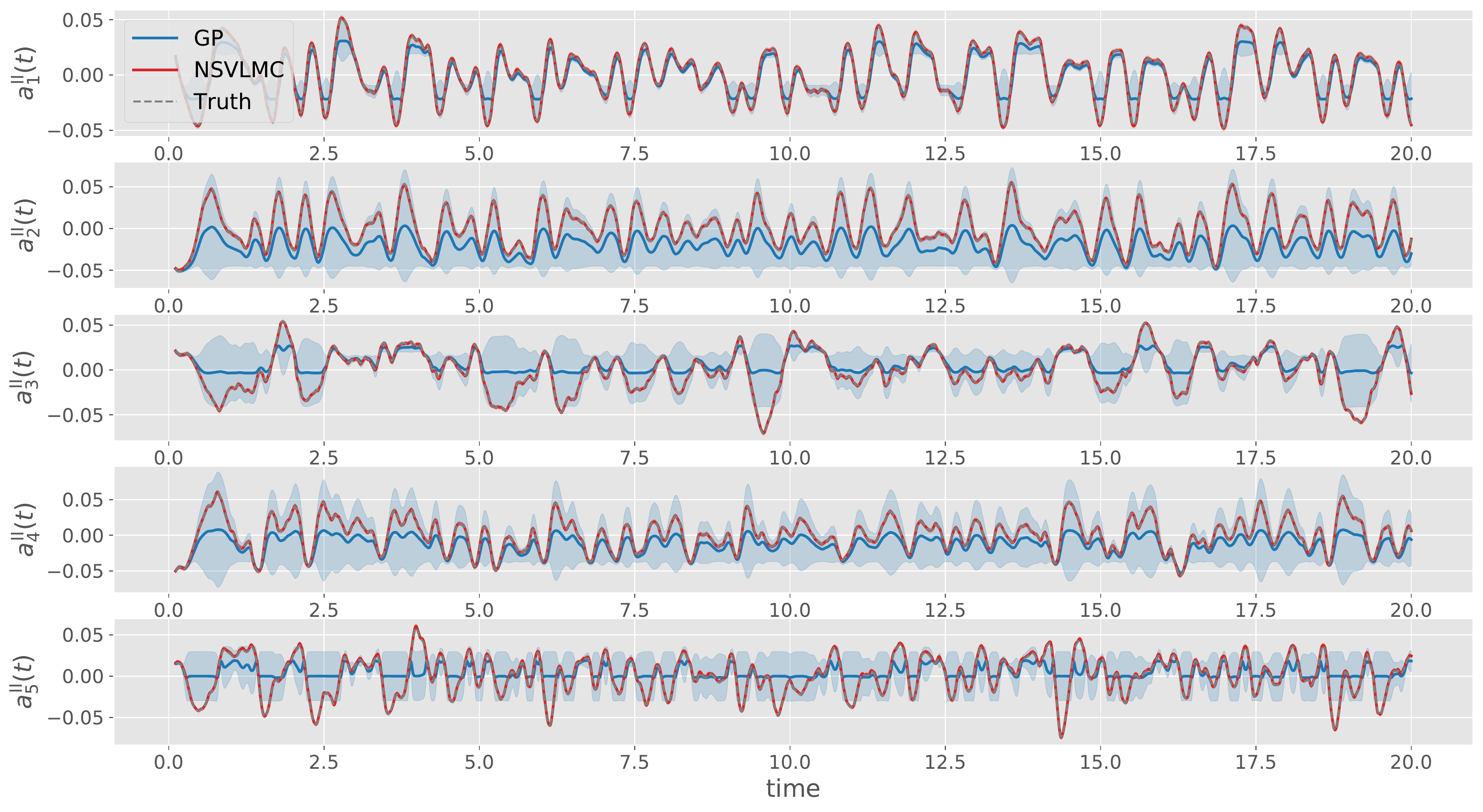}
	\caption{The five time series coefficients of case II of the fluidized bed predicted by GP and NSVLMC, respectively. The shaded regions indicate 95\% confidence interval of the prediction.}
	\label{fig_fluid_bed_gp_nsvlmc}
\end{figure}

\begin{figure}[t!]
	\centering
	\includegraphics[width=.5\textwidth]{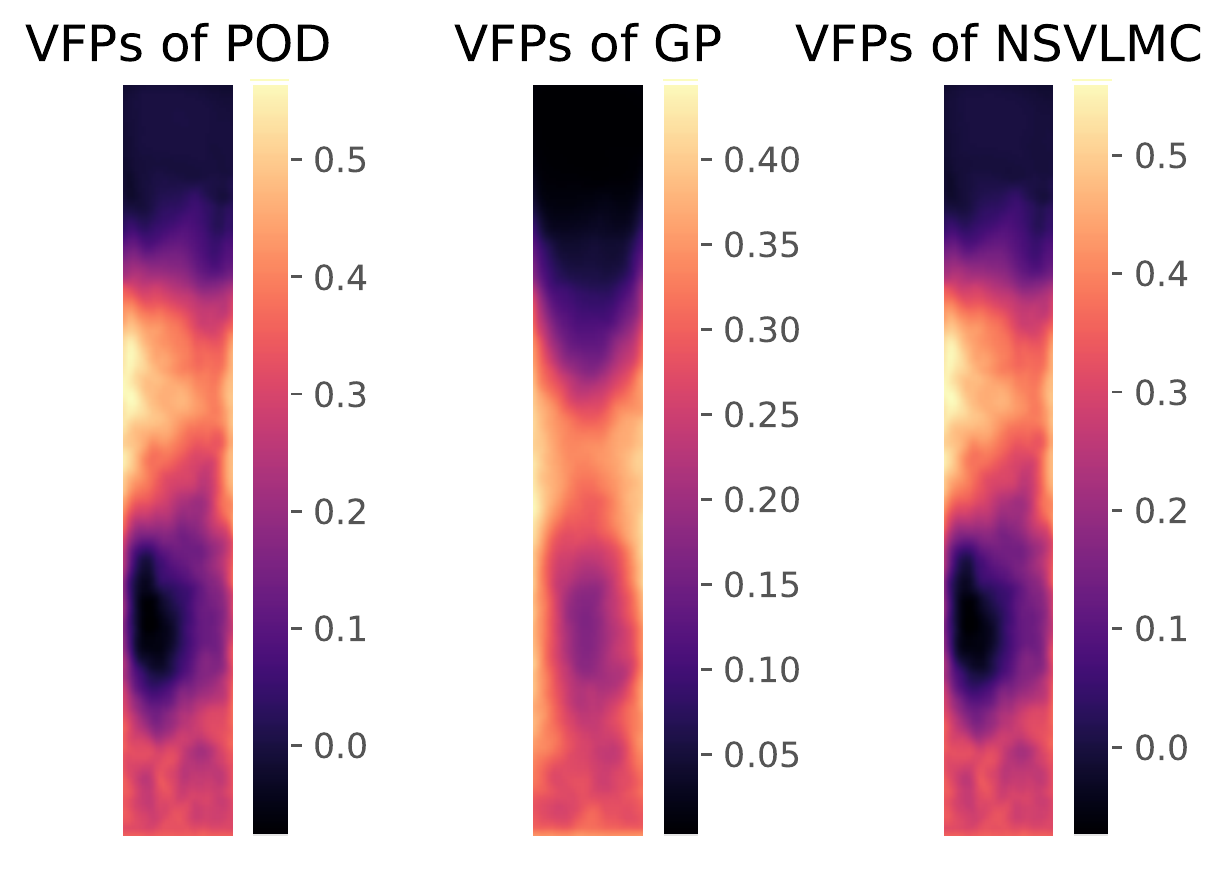}
	\caption{The VFPs of fluidized bed recovered by the five-order POD, the GP and the proposed NSVLMC, respectively, at $t=0.6$s.}
	\label{fig_VFPs}
\end{figure}

Table~\ref{tab_fluid_bed} summarizes the comparative results of NSVLMC against other neural embeddings, the original SVLMC and the baseline GP to predict the time series coefficients of case II of the fluidized bed.\footnote{Note that since GP is independent of random seed, it produces the same results for multiple runs.} It is found that all the LMCs except the flexible NGPRN outperform the single-task GP for modeling the five time series coefficients of case II by leveraging knowledge shared from case I, see the illustration of NSVLMC versus GP in Fig.~\ref{fig_fluid_bed_gp_nsvlmc}. Besides, we can again observe the superiority of our NSVLMC in comparison to other neural embeddings, and the results of flexible NGPRN here again reveal the serious issue of poor generalization. Finally, after predicting the time series coefficients of case II, we can use the POD expression~\eqref{eq_pod} to recover the VFPs, as shown in Fig.~\ref{fig_VFPs}. It is observed that in comparison to the VFPs recovered by GP, the VFPs recovered by NSVLMC at $t=0.6$s are closer to the VFPs of POD due to the high quality of time series coefficients predictions. Note that the quality of VFP recovery could be further improved by POD and NSVLMC using higher orders.

\section{Conclusions} \label{sec_conclusion}
In order to improve the performance of LMC, this paper develops a new LMC paradigm wherein the neural embedding is adopted to induce rich yet diverse latent GPs in a high-dimensional latent space, and the scalable yet tight ELBO has been derived for high quality of model inference. We compare the proposed NSVLMC against existing LMCs on various multi-task learning cases to showcase its methodological characteristics and superiority. Further improvements related to the NSVLMC model could be the extensions on handling heterogeneous input domains (e.g., the input domains of tasks may have different dimensions)~\cite{hebbal2021multi, mao2021multiview} and massive related outputs (e.g., the direct modeling of fluids)~\cite{zhe2019scalable, wang2020multi}.

\section*{Acknowledgements}
This work was supported by the National Natural Science Foundation of China (52005074), and the Fundamental Research Funds for the Central Universities (DUT19RC(3)070). Besides, it was partially supported by the Research and Innovation in Science and Technology Major Project of Liaoning Province (2019JH1-10100024), and the MIIT Marine Welfare Project (Z135060009002).

\section*{Appendix}
\begin{appendices}
\section{Expressions for the components in ELBO~\eqref{eq_elbo_tight}} \label{app_elbo}
For the inner term $\tilde{\mathcal{L}}_{\mbf{AB}}$ in the expectation $\mathbb{E}_{q(\mbf{A})} \left[ \log \mathbb{E}_{p(\mbf{B})}\left[ \exp(\tilde{\mathcal{L}}_{\mbf{AB}}) \right] \right]$ of ELBO~\eqref{eq_elbo_tight}, we have
\begin{align} \label{eq_elbo_fact}
\begin{aligned}
\tilde{\mathcal{L}}_{\mbf{AB}} =& \mathbb{E}_{q(\mbf{f})} [\log p(\mbf{y}|\mbf{f}, \mbf{A}, \mbf{B})] \\
=& \left\langle \log p(\mbf{y}| \mbf{f}, \mbf{A}, \mbf{B}) \right\rangle_{\prod_{q=1}^Q q(\mbf{f}_q)} \\
=& \sum_{c=1}^C \sum_{i=1}^{N^c} \mathbb{E}_{\prod_{q=1}^Q q(f_{q,i}^c)} \left[\log \mathcal{N}\left(y_i^c \left| \sum_{h=1}^H a_h^c \sum_{q=1}^Q b_q^h(\mbf{x}_i^c) f_q(\mbf{x}_i^c), \nu^c_{\epsilon} \right.\right) \right] \\
=& \sum_{c=1}^C \sum_{i=1}^{N^c} \log \mathcal{N}\left(y_i^c \left| \sum_{h=1}^H \sum_{q=1}^Q a_h^c b_q^h(\mbf{x}_i^c) \mu_{q,i}^c, \nu^c_{\epsilon} \right.\right) \\
&- \frac{1}{2\nu^c_{\epsilon}} \left(\sum_{h=1}^H \sum_{q=1}^Q (a_h^c b_q^h(\mbf{x}_i^c))^2 \nu_{q,i}^c \right),
\end{aligned}
\end{align}
where the individual mean and variance express respectively as
\begin{align}
	\mu_{q,i}^c &= \bm{\mu}_{q}(\mbf{x}_i^c) = k_q(\mbf{x}_i^c,\mbf{Z}_q) \mbf{K}^{-1}_{Z_q} \mbf{m}_q, \\
	\nu_{q,i}^c &= \bm{\Sigma}_{q}(\mbf{x}_i^c) = k_q(\mbf{x}_i^c,\mbf{x}_i^c) + k_q(\mbf{x}_i^c,\mbf{Z}_q) \mbf{K}^{-1}_{Z_q} [\mbf{S}_q \mbf{K}^{-1}_{Z_q} - \mbf{I}] k_q^{\mathsf{T}}(\mbf{x}_i^c,\mbf{Z}_q).
\end{align}
Thereafter, the unbiased evaluation of the expectation could be conducted through the Markov Chain Monte Carlo (MCMC) sampling method by the samples from the variational Gaussian posterior $q(\mbf{A})$ and the Gaussian prior $p(\mbf{B})$. It is found that the factorized expression over both data points and tasks in~\eqref{eq_elbo_fact} makes the ELBO be efficiently evaluated and optimized through the mini-batch fashion and stochastic optimizer, for example, Adam~\cite{kingma2015adam}.

For the remaining two KL terms, due to the Gaussian form, all of them could be calculated analytically as
\begin{align}
\begin{aligned}
\mathrm{KL}[q(\mbf{A})||p(\mbf{A})] =& \sum_{c=1}^C \sum_{h=1}^H \mathrm{KL}[q(a_h^c)||p(a_h^c)] \\
=& \frac{1}{2} \left(-\log |\mathrm{diag}(\bm{\nu}_{\mbf{A}})| - C \times H + \mathrm{Tr}(\mathrm{diag}(\bm{\nu}_{\mbf{A}})) + \bm{\mu}_{\mbf{A}}^{\mathsf{T}} \bm{\mu}_{\mbf{A}} \right),
\end{aligned} 
\end{align}
and
\begin{align}
\begin{aligned}
\mathrm{KL}[q(\mbf{u})||p(\mbf{u})] =& \sum_{q=1}^Q \mathrm{KL}[q(\mbf{u}_q)||p(\mbf{u}_q)] \\
=& \sum_{q=1}^Q \frac{1}{2} \left(\log \frac{|\mbf{K}_{Z_q}|}{|\mbf{S}_q|} -N + \mathrm{Tr}(\mbf{K}_{Z_q}^{-1} \mbf{S}_q) + \mbf{m}_q^{\mathsf{T}} \mbf{K}_{Z_q}^{-1} \mbf{m}_q \right).
\end{aligned}
\end{align}

\section{Experimental configurations} \label{sec_exp_details}
The experimental configurations for the numerical cases in our comparative study in section~\ref{sec_exp} are detailed as following.

As for data preprocessing, we normalize the inputs along each dimension to have zero mean and unit variance, and specifically, this normalization has been applied for the outputs to fulfill the GP model assumption. Besides, we have ten runs of model training with different random seeds on each case to quantify the algorithmic robustness.

For the proposed NSVLMC model, it adopts the NN prior for the mean and variance of $p(\mbf{B})$ in~\eqref{eq_p_B} as
\begin{align}
\bm{\mu}_{\mbf{B}} &= \mbf{W}_{\mu}^{\mathsf{T}} \mathrm{MLP}_{\bm{\theta}}(\mbf{x}) + \mbf{b}_{\mu}, \\
\bm{\nu}_{\mbf{B}} &= \nu_0 \times \mathtt{Sigmoid}(\mbf{W}_{\nu}^{\mathsf{T}} \mathrm{MLP}_{\bm{\theta}}(\mbf{x}) + \mbf{b}_{\nu}),
\end{align} 
where the base $\mathrm{MLP}_{\bm{\theta}}(.)$ accepts the $D$-dimensional input $\mbf{x}$ and has three fully-connected (FC) hidden layers ``FC($Q\times H$)-FC($Q\times H$)-FC($Q\times H$)'', each of which employs the tanh activation function and has $Q\times H$ hidden units with the weights initialized through the Xavier method~\cite{glorot2010understanding} and the biases initialized as zeros; the weights $\mbf{W}_{\mu}^{\mathsf{T}}$ and biases $\mbf{b}_{\mu}$ are the parameters for the final FC layer outputting the mean $\bm{\mu}_{\mbf{B}}$; similarly, the weights $\mbf{W}_{\nu}^{\mathsf{T}}$ and biases $\mbf{b}_{\nu}$ are the parameters for the final FC layer outputting the variance $\bm{\nu}_{\mbf{B}}$ with the sigmoid activate function which makes the output be within $[0,1]$; finally, the small positive parameter $\nu_0$ scales the variance, and it is initialized as $10^{-4}$ in order to yield nearly deterministic behaviors at the beginning for speeding up model training. 

As for the GP part of NSVLMC, we employ the SE kernel~\eqref{eq_SE} with (i) the length-scales $\{l_i \}_{i=1}^D$ initialized as 0.1 for the toy case, the \texttt{Jura} and \texttt{EEG} datasets, and the fluidized bed case, and they are initialized as 0.5 for the \texttt{Sarcos} dataset; and (ii) the output scale $\sigma^2_f$ is initialized as 1.0. We use the same inducing size $M_q = M$ ($1 \le q \le Q$) for latent GPs. Specifically, for the small-scale cases (\texttt{Jura}, \texttt{EEG} and case A of \texttt{Sarcos}), we use all the training inputs to initialize the inducing positions due to the low model complexity; while for the large-scale cases, we have $M=100$ and initialize them though the $k$-means clustering technique from the \texttt{scikit}-\texttt{learn} package~\cite{pedregosa2011scikit} for cases B and C of the \texttt{Sarcos} dataset, and we adopt $M=500$ for the fluidized bed case. The ELBO~\eqref{eq_elbo_iwvi} in training is estimated through $S=10$ samples; while for predicting, we sample 100 points from $p(\mbf{y}_*|\mbf{y})$ in~\eqref{eq_y*}.

As for the parameters $Q$ and $H$ in NSVLMC, we have $Q=2$ for all the cases except the \texttt{EEG} dataset which adopts $Q=4$; we have $H=20$ for the \texttt{Jura} and \texttt{EEG} datasets, and $H=10$ for cases A and B of the \texttt{Sarcos} dataset but $H=100$ for case C due to the low task correlation, and finally we have $H=10$ for the fluidized bed case.

As for the optimization, we employ the well-known Adam optimizer~\cite{kingma2015adam} with the learning rate of $5\times10^{-3}$,\footnote{Since the proposed NSVLMC is a hybrid model of NN and GP, we adopt a mild learning rate $5\times10^{-3}$ according to the suggestion in~\cite{liu2021deep}.} and run it over 10000 iterations for all the cases except the \texttt{Sarcos} dataset which runs up to 20000 iterations. The mini-batch size $|\mathcal{B}|$ takes 32 for the \texttt{Jura} and \texttt{Sarcos} datasets and the fluidized case, and it is 64 for the \texttt{EEG} dataset.

\section{Error criteria for model evaluation} \label{sec_err_criteria}
The expressions of the error criteria employed in the comparison study in section~\ref{sec_exp} for model evaluation are elaborated respectively as below.

First, for the $c$-th output, given $N^c_*$ test points $\{\mbf{X}^c_* \in \mathbb{R}^{N^c_* \times D}, \mbf{y}^c_* \in \mathbb{R}^{N^c_*}\}$, the MAE criterion is used to quantify the precision of prediction mean as
\begin{align}
e^c_{\mathrm{MAE}} = \frac{1}{N^c_*} \sum_{i=1}^{N^c_*} |\mu^c_{*i} - y^c_{*i}|,
\end{align}
where $\mu^c_{*i}$ is the prediction mean at test point $\mbf{x}^c_i$, and $y^c_{*i}$ is the true observation at $\mbf{x}^c_i$. Second, different from MAE, the SMSE is a normalized error criterion expressed as
\begin{align}
e^c_{\mathrm{SMSE}} = \frac{1}{N^c_*} \frac{\sum_{i=1}^{N^c_*} (\mu^c_{*i} - y^c_{*i})^2}{\mathrm{var}(\mbf{y}^c)},
\end{align}
where $\mathrm{var}(\mbf{y}^c)$ is the estimated variance of training outputs of the $c$-th task. It is noted that the SMSE equals to one when the model always predict the mean of $\mbf{y}^c$. Third, different from both MAE and SMSE, the informative NLL criteria is employed to further quantify the quality of predictive distribution. It is thus expressed as
\begin{align}
e^c_{\mathrm{NLL}} = \frac{1}{N^c_*} \sum_{i=1}^{N^c_*} \frac{1}{2} \left[ \frac{(\mu^c_{*i} - y^c_{*i})^2}{\nu^c_{*i}} + \log(2\pi \nu^c_{*i})\right],
\end{align}
where $\nu^c_{*i}$ is the prediction variance at test point $\mbf{x}^c_i$. For all the three criteria, lower is better.

\end{appendices}

\section*{References}
\bibliography{MTGP}

\end{document}